# On Quadratic Convergence of DC Proximal Newton Algorithm for Nonconvex Sparse Learning in High Dimensions

Xingguo Li, Lin F. Yang, Jason Ge, Jarvis Haupt, Tong Zhang, and Tuo Zhao*

February 15, 2018

## Abstract

We propose a DC proximal Newton algorithm for solving nonconvex regularized sparse learning problems in high dimensions. Our proposed algorithm integrates the proximal Newton algorithm with multi-stage convex relaxation based on difference of convex (DC) programming, and enjoys both strong computational and statistical guarantees. Specifically, by leveraging a sophisticated characterization of sparse modeling structures/assumptions (i.e., local restricted strong convexity and Hessian smoothness), we prove that within each stage of convex relaxation, our proposed algorithm achieves (local) quadratic convergence, and eventually obtains a sparse approximate local optimum with optimal statistical properties after only a few convex relaxations. Numerical experiments are provided to support our theory.

## 1 Introduction

We consider a high dimensional regression or classification problem: Given $n$ independent observations $\{x_i, y_i\}_{i=1}^n \subset \mathbb{R}^d \times \mathbb{R}$ sampled from a joint distribution $\mathcal{D}(X, Y)$, we are interested in learning the conditional distribution $\mathbb{P}(Y|X)$ from the data. A popular modeling approach is the Generalized Linear Model (GLM) (McCullagh, 1984), which assumes

$$\mathbb{P}(Y|X;\theta^*) \propto \exp\left(\frac{YX^\top \theta^* - \psi(X^\top \theta^*)}{c(\sigma)}\right),$$

where $c(\sigma)$ is a scaling parameter, and $\psi$ is the cumulant function. A natural approach to estimate $\theta^*$ is the Maximum Likelihood Estimation (MLE) (Pfanzagl, 1994), which essentially minimizes the negative log-likelihood of the data given parameters. However, MLE often performs poorly in parameter estimation in high dimensions due to the curse of dimensionality (Bühlmann and Van De Geer, 2011).

---
*Xingguo Li and Jarvis Haupt are affiliated with Department of Electrical and Computer Engineering at University of Minnesota; Lin Yang is affiliated with Department of Computer Science and Physics at Johns Hopkins University; Jason Ge is affiliated with Department of Operations Research and Financial Engineering at Princeton University; Tong Zhang is affiliated with Tecent AI Lab; Tuo Zhao is affiliated with School of Industrial and Systems Engineering at Georgia Institute of Technology. The authors acknowledge support from DARPA YFA N66001-14-1-4047, NSF Grant IIS-1447639, and Doctoral Dissertation Fellowship from University of Minnesota. Correspondence to: Xingguo Li <lixx1661@umn.edu> and Tuo Zhao <tuo.zhao@isye.gatech.edu>.

To address this issue, machine learning researchers and statisticians follow Occam's razor principle, and propose sparse modeling approaches (Tibshirani, 1996; van de Geer, 2008; Raginsky et al., 2010; Belloni et al., 2011). These sparse modeling approaches assume that $\theta^*$ is a sparse vector with only $s^*$ non-zero entries, where $s^* < n \ll d$. This implies that only a few variables in $X$ are essentially relevant to modeling, which is actually very natural to many real world applications, such as genomics and medical imaging (Neale et al., 2012; Eloyan et al., 2012; Liu et al., 2015). Numerous empirical results have corroborated the success of sparse modeling in high dimensions. Sparse modeling approaches usually obtain a sparse estimator of $\theta^*$ by solving the following regularized optimization problem,

$$\overline{\theta} = \underset{\theta \in \mathbb{R}^d}{\operatorname{argmin}} \mathcal{L}(\theta) + \mathcal{R}_{\lambda_{\text{tgt}}}(\theta), \tag{1.1}$$

where $\mathcal{L} : \mathbb{R}^d \to \mathbb{R}$ is a twice differentiable convex loss function (e.g., negative log-likelihood or pseudo-likelihood), $\mathcal{R}_{\lambda_{\text{tgt}}} : \mathbb{R}^d \to \mathbb{R}$ is a sparsity-inducing decomposable regularizer, i.e., $\mathcal{R}_{\lambda_{\text{tgt}}}(\theta) = \sum_{j=1}^{d} r_{\lambda_{\text{tgt}}}(\theta_j)$ with $r_{\lambda_{\text{tgt}}} : \mathbb{R} \to \mathbb{R}$, and $\lambda_{\text{tgt}} > 0$ is the regularization parameter. Most of the existing sparse modeling approaches can be cast as special examples of (1.1), such as sparse linear regression Tibshirani (1996), sparse logistic regression (van de Geer, 2008), and sparse Poisson regression (Raginsky et al., 2010).

For convex regularizers, e.g., $\mathcal{R}_{\text{tgt}}(\theta) = \lambda_{\text{tgt}} \|\theta\|_1$ (Tibshirani, 1996), we can obtain global optima in polynomial time and characterize their statistical properties. However, convex regularizers incur large estimation bias, since they induces too large penalty for the coefficients with large magnitudes. To address this issue, several nonconvex regularizers are proposed, including the minimax concave penalty (MCP, Zhang (2010a)), smooth clipped absolute deviation (SCAD, Fan and Li (2001)), and capped $\ell_1$-regularization (Zhang, 2010b). The obtained estimator (e.g., hypothetical global optima to (1.1)) can achieve faster statistical rates of convergence than their convex counterparts in parameter estimation (Negahban et al., 2012; Loh and Wainwright, 2015; Wang et al., 2014; Fan et al., 2015).

**Related Work:** Despite of these superior statistical guarantees, nonconvex regularizers raise greater computational challenge than convex regularizers in high dimensions. Popular iterative algorithms for convex optimization, such as proximal gradient descent (Beck and Teboulle, 2009; Nesterov, 2013) and coordinate descent (Luo and Tseng, 1992; Shalev-Shwartz and Tewari, 2011), no longer have strong global convergence guarantees for nonconvex optimization. Therefore, establishing statistical properties of the estimators obtained by these algorithms becomes very challenging, which explains why existing theoretical studies on computational and statistical guarantees for nonconvex regularized sparse modeling approaches are so limited until recent rise of a new area named "statistical optimization". Specifically, machine learning researchers start to incorporate certain structures of sparse modeling (e.g. restricted strong convexity, large regularization effect) into the algorithmic design and convergence analysis for nonconvex optimization.

This further motivates a few recent progresses: Loh and Wainwright (2015) propose proximal gradient algorithms for a family of nonconvex regularized estimators with a linear convergence to an approximate local optimum with suboptimal statistical guarantees; Wang et al. (2014); Zhao et al. (2014) further propose homotopy proximal gradient and coordinate gradient descent algorithms with a linear convergence to a local optimum with optimal statistical guarantees; Zhang (2010b); Fan et al. (2015) propose a multistage convex relaxation based (also known as Difference of Convex (DC) Programming) proximal gradient algorithm, which can guarantee an approximate local optimum with optimal statistical properties. The computational analysis in Fan et al. (2015) further shows that within each stage of the convex relaxation, the proximal gradient algorithm achieves a (local) linear convergence to a unique sparse global optimum for the relaxed convex subproblem.

**Motivation:** The aforementioned approaches only consider first order algorithms, such as proximal gradient descent and proximal coordinate gradient descent. The second order algorithms with theoretical guarantees are still largely missing for high dimensional nonconvex regularized sparse modeling approaches, but this does not suppress the enthusiasm of applying heuristic second order algorithms to real world problems. Some evidences have already corroborated their superior computational performance over first order algorithms (e.g. `glmnet`, Friedman et al. (2010)). This further motivates our attempt towards understanding the second order algorithms in high dimensions.

**Our Contribution:** We study a multistage convex relaxation based proximal Newton algorithm for nonconvex regularized sparse learning. This algorithm is not only highly efficient in practice, but also enjoys strong computational and statistical guarantees in theory. Specifically, by leveraging a sophisticated characterization of local restricted strong convexity and Hessian smoothness, we prove that within each stage of convex relaxation, our proposed algorithm maintains the solution sparsity, and achieves a (local) quadratic convergence, which is a significant improvement over the (local) linear convergence of the proximal gradient algorithm in Fan et al. (2015) (See more details in later sections). This eventually allows us to obtain an approximate local optimum with optimal statistical properties after only a few number of convex relaxation stages. Numerical experiments are provided to support our theory. To the best of our knowledge, this is the first of second order based approaches for high dimensional sparse learning using convex/nonconvex regularizers with strong statistical and computational guarantees.

**Organization:** The rest of this paper is as follows: In Section 2, we introduce the basic assumptions of the objective function and our algorithm; In Section 3, we present both statistical and computational theories that guarantee the convergence of our proposed algorithm; In Section 4, we provide numerical experiments to support our theories; In Section 5, we detailedly explain why our second order algorithm is superior to the existing first order algorithms in practice, and discuss the extensions of our methodology and theory to proximal sub-sampled Newton and Quasi-Newton algorithms; The proof sketches of our theories are presented in Section 6; The technical lemmas and supplementary materials are presented in Appendix.

**Notations**: Given a vector $v \in \mathbb{R}^d$, we denote the set of index for non-zero entries as $\text{supp}(v)$, the number of non-zero entries as $\|v\|_0 = \sum_j \mathbb{1}(v_j \neq 0)$, the $p$-norm as $\|v\|_p = (\sum_{j=1}^d |v_j|^p)^{1/p}$ for a real $p > 0$, $\|v\|_\infty = \max_j |v_j|$, and the subvector with the $j$-th entry removed as $v_{\backslash j} = (v_1, \ldots, v_{j-1}, v_{j+1}, \ldots, v_d)^\top \in \mathbb{R}^{d-1}$. Given an index set $\mathcal{A} \subseteq \{1, \ldots, d\}$, $\mathcal{A}_\perp = \{j \mid j \in \{1, \ldots, d\}, j \notin \mathcal{A}\}$ is the complementary set to $\mathcal{A}$. We use $v_\mathcal{A}$ to denote a subvector of $v$ indexed by $\mathcal{A}$. Given a matrix $A \in \mathbb{R}^{d \times d}$, we use $A_{*j}$ ($A_{k*}$) to denote the $j$-th column ($k$-th row) and $\Lambda_{\max}(A)$ ($\Lambda_{\min}(A)$) as the largest (smallest) eigenvalue of $A$. We define $\|A\|_F^2 = \sum_j \|A_{*j}\|_2^2$ and $\|A\|_2 = \sqrt{\Lambda_{\max}(A^\top A)}$. We denote $A_{\backslash i \backslash j}$ as the submatrix of $A$ with the $i$-th row and the $j$-th column removed, $A_{\backslash ij}$ ($A_{i \backslash j}$) as the $j$-th column ($i$-th row) of $A$ with its $i$-th ($j$-th) entry removed, and $A_{\mathcal{A}\mathcal{A}}$ as a submatrix of $A$ with both row and column indexed by $\mathcal{A}$. If $A$ is a positive semidefinite matrix, we define $\|v\|_A = \sqrt{v^\top A v}$ as the induced seminorm for vector $v$. We use conventional notation $\mathcal{O}(\cdot), \Omega(\cdot), \Theta(\cdot)$ to denote the limiting behavior, ignoring constant, and $\mathcal{O}_P(\cdot)$ to denote the limiting behavior in probability. $C_1, C_2, \ldots$ are denoted as generic positive constants.

## 2 DC Proximal Newton Algorithm

Throughout the rest of the paper, we assume: (1) $\mathcal{L}(\theta)$ is nonstrongly convex and twice continuously differentiable, e.g., the negative of log-likelihood function for the generalized linear model (GLM); (2) $\mathcal{L}(\theta)$ takes an additive form, i.e.,

$$\mathcal{L}(\theta) = \frac{1}{n} \sum_{i=1}^n \ell_i(\theta),$$

where each $\ell_i(\theta)$ is associated with an observation $(x_i, y_i)$ for $i = 1, \ldots, n$. Take GLM as an example, we have

$$\ell_i(\theta) = \psi(x_i^\top \theta) - y_i x_i^\top \theta,$$

where $\psi$ is the cumulant function.

For nonconvex regularization, we use the capped $\ell_1$ regularizer (Zhang, 2010b) defined as

$$\mathcal{R}_{\lambda_{\text{tgt}}}(\theta) = \sum_{j=1}^d r_{\text{tgt}}(\theta_j) = \lambda_{\text{tgt}} \sum_{j=1}^d \min\{|\theta_j|, \beta \lambda_{\text{tgt}}\}, \tag{2.1}$$

where $\beta > 0$ is an additional tuning parameter[1]. Our algorithm and theory can also be extended to the SCAD and MCP regularizers in a straightforward manner (Zhang, 2010a; Fan and Li, 2001). As shown in Figure 1, $r_{\lambda_{\text{tgt}}}(\theta_j)$ can be decomposed as the difference of two convex functions (Boyd and Vandenberghe, 2009),

$$r_\lambda(\theta_j) = \underbrace{\lambda|\theta_j|}_{\text{convex}} - \underbrace{\max\{\lambda|\theta_j| - \beta\lambda^2, 0\}}_{\text{convex}}.$$

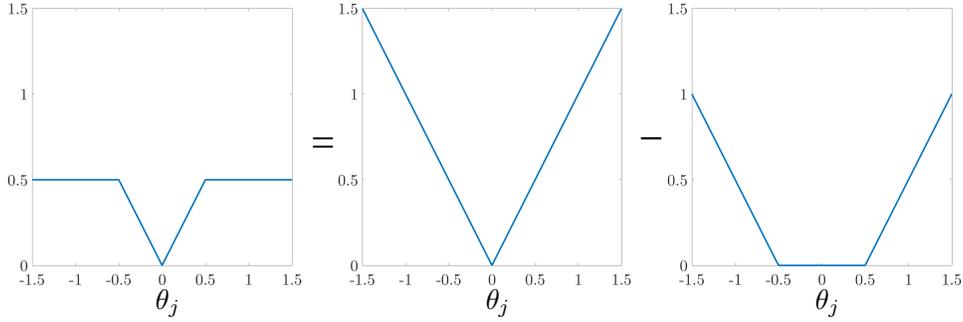

Figure 1: The capped $\ell_1$ regularizer is the difference of two convex functions. This allows us to relax the nonconvex regularizer based the concave duality.

This motivates us to apply the difference of convex (DC) programming approach to solve the nonconvex problem. We then introduce the DC proximal Newton algorithm, which contains three components: the multistage convex relaxation, warm initialization, and proximal Newton algorithm.

**(I) The multistage convex relaxation** is essentially a sequential optimization framework (Zhang, 2010b)[2]. At the $\{K+1\}$-th stage, we have the output solution from the previous stage $\widehat{\theta}^{\{K\}}$. For notational simplicity, for all $j = 1, \ldots, d$, we define a regularization vector $\lambda^{\{K+1\}} \in \mathbb{R}^d$ as

$$\lambda^{\{K+1\}} = \left(\lambda_1^{\{K+1\}}, \ldots, \lambda_d^{\{K+1\}}\right)^\top, \quad \text{where} \quad \lambda_j^{\{K+1\}} = \lambda_{\text{tgt}} \cdot \mathbb{1}\left(|\widehat{\theta}_j^{\{K\}}| \le \beta \lambda_{\text{tgt}}\right).$$

Let $\odot$ be the Hadamard (entrywise) product. We solve a convex relaxation of (1.1) at $\theta = \widehat{\theta}^{\{K\}}$ as follows,

$$\overline{\theta}^{\{K+1\}} = \operatorname*{argmin}_{\theta \in \mathbb{R}^d} \mathcal{F}_{\lambda^{\{K+1\}}}(\theta), \text{ where } \mathcal{F}_{\lambda^{\{K+1\}}}(\theta) = \mathcal{L}(\theta) + \|\lambda^{\{K+1\}} \odot \theta\|_1, \tag{2.2}$$

where $\|\lambda^{\{K+1\}} \odot \theta\|_1 = \sum_{j=1}^d \lambda_j^{\{K+1\}} |\theta_j|$. One can verify that $\|\lambda^{\{K+1\}} \odot \theta\|_1$ is essentially a convex relaxation of $\mathcal{R}_{\lambda_{\text{tgt}}}(\theta)$ at $\theta = \widehat{\theta}^{\{K\}}$ based on the concave duality in DC programming.

**Remark 2.1.** *We emphasize that $\overline{\theta}^{\{K\}}$ denotes the unique sparse global optimum for (2.2) (The uniqueness will be elaborated in later sections), and $\widehat{\theta}^{\{K\}}$ denotes the output solution for (2.2) when we terminate the iteration at the K-th convex relaxation stage. The stopping criterion will be explained later.*

**(II) The warm initialization** is the first stage of DC programming, which solves the $\ell_1$-regularized counterpart of (1.1),

$$\overline{\theta}^{\{1\}} = \operatorname*{argmin}_{\theta \in \mathbb{R}^d} \mathcal{L}(\theta) + \lambda_{\text{tgt}} \|\theta\|_1. \tag{2.3}$$

---

[1] The capped $\ell_1$ regularizer is also independently proposed by Shen et al. (2012) with a different name – "Truncated $\ell_1$ Regularizer".

[2] The DC programming approach is also independently proposed by Shen et al. (2012) as heuristics, and their statistical theory is still based on the hypothetical global optima.

This is an intuitive choice for sparse statistical recovery, since the $\ell_1$-regularized estimator can give us a good initialization, which is sufficiently close to $\theta^*$. Note that (2.3) equivalent to (2.2) with $\lambda_j^{\{1\}} = \lambda_{\text{tgt}}$ for all $j = 1, \ldots, d$, which can be viewed as the convex relaxation of (1.1) by taking $\widehat{\theta}^{\{0\}} = \mathbf{0}$ for the first stage.

**(III) The proximal Newton algorithm** proposed in Lee et al. (2014) is then applied to solve the convex subproblem (2.2) at each stage, including the warm initialization (2.3). For notational simplicity, we omit the stage index $\{K\}$ for all intermediate updates of $\theta$, and only use $(t)$ as the iteration index within the $K$-th stage for all $K \geq 1$. Specifically, at the $K$-th stage, given $\theta^{(t)}$ at the $t$-th iteration of the proximal Newton algorithm, we consider a quadratic approximation of (2.2) at $\theta^{(t)}$ as follows,

$$\mathcal{Q}(\theta; \theta^{(t)}, \lambda^{\{K\}}) = \mathcal{L}(\theta^{(t)}) + (\theta - \theta^{(t)})^\top \nabla \mathcal{L}(\theta^{(t)}) + \frac{1}{2} \|\theta - \theta^{(t)}\|^2_{\nabla^2 \mathcal{L}(\theta^{(t)})} + \|\lambda^{\{K\}} \odot \theta\|_1, \tag{2.4}$$

where $\|\theta - \theta^{(t)}\|^2_{\nabla^2 \mathcal{L}(\theta^{(t)})} = (\theta - \theta^{(t)})^\top \nabla^2 \mathcal{L}(\theta^{(t)})(\theta - \theta^{(t)})$. We then take

$$\theta^{(t+\frac{1}{2})} = \underset{\theta}{\operatorname{argmin}} \ \mathcal{Q}(\theta; \theta^{(t)}, \lambda^{\{K\}}).$$

Since $\mathcal{L}(\theta) = \frac{1}{n} \sum_{i=1}^n \ell_i(\theta)$ takes an additive form, we can avoid directly computing the $d$ by $d$ Hessian matrix in (2.4). Alternatively, in order to reduce the memory usage when $d$ is large, we rewrite (2.4) as a regularized weighted least square problem as follows

$$\mathcal{Q}(\theta; \theta^{(t)}, \lambda^{\{K\}}) = \frac{1}{n} \sum_{i=1}^n w_i (z_i - x_i^\top \theta)^2 + \|\lambda^{\{K\}} \odot \theta\|_1 + \text{constant}, \tag{2.5}$$

where $w_i$'s and $z_i$'s are some easy to compute constants depending on $\theta^{(t)}$, $\ell_i(\theta^{(t)})$'s, $x_i$'s, and $y_i$'s.

**Remark 2.2.** *Existing literature has shown that the $\ell_1$-regularized quadratic problem in (2.5) can be efficiently solved by coordinate descent algorithms in conjunction with the active set strategy (Zhao et al., 2014). See more details in Friedman et al. (2010) and Appendix A.*

For the first stage (i.e., warm initialization), we require an additional backtracking line search procedure to guarantee the descent of the objective value (Lee et al., 2014). Specifically, we denote

$$\Delta \theta^{(t)} = \theta^{(t+\frac{1}{2})} - \theta^{(t)}.$$

Then we start from $\eta_t = 1$ and use a backtracking line search procedure to find the optimal $\eta_t \in (0, 1]$ such that the Armijo condition (Armijo, 1966) holds. Specifically, given a constant $\mu \in (0.9, 1)$, we update $\eta_t = \mu^q$ from $q = 0$ and find the smallest nonnegative integer $q$ such that

$$\mathcal{F}_{\lambda^{\{1\}}}(\theta^{(t)} + \eta_t \Delta \theta^{(t)}) \leq \mathcal{F}_{\lambda^{\{1\}}}(\theta^{(t)}) + \alpha \eta_t \gamma_t,$$

where $\alpha \in (0, \frac{1}{2})$ is a fixed constant and

$$\gamma_t = \nabla \mathcal{L}\left(\theta^{(t)}\right)^\top \cdot \Delta \theta^{(t)} + \|\lambda^{\{1\}} \odot \left(\theta^{(t)} + \Delta \theta^{(t)}\right)\|_1 - \|\lambda^{\{1\}} \odot \theta^{(t)}\|_1.$$

We then set $\theta^{(t+1)}$ as $\theta^{(t+1)} = \theta^{(t)} + \eta_t \Delta\theta^{(t)}$ and terminate the iterations for the smallest $t$ when the following approximate KKT condition holds:

$$\omega_{\lambda^{\{1\}}}\left(\theta^{(t)}\right) = \min_{\xi \in \partial\|\theta^{(t)}\|_1} \|\nabla\mathcal{L}(\theta^{(t)}) + \lambda^{\{1\}} \odot \xi\|_\infty \leq \varepsilon,$$

where $\varepsilon$ is a predefined precision parameter. Then we set the output solution as $\widehat{\theta}^{\{1\}} = \theta^{(t)}$. Note that $\widehat{\theta}^{\{1\}}$ is then used as the initial solution for the second stage of convex relaxation (2.2). The proximal Newton algorithm with backtracking line search is summarized in Algorithm 1.

Such a backtracking line search procedure is not necessary at $K$-th stage for all $K \geq 2$. In other words, we simply take $\eta_t = 1$ and $\theta^{(t+1)} = \theta^{(t)} + \Delta\theta^{(t)} = \theta^{(t+\frac{1}{2})}$ for all $t \geq 0$ when $K \geq 2$. This leads to more efficient updates for the proximal Newton algorithm from the second stage of convex relaxation (2.2). We summarize our proposed DC proximal Newton algorithm in Algorithm 2.

---

**Algorithm 1** Proximal Newton Algorithm (ProxNewton)

**Input:** $\theta^{(0)}, \lambda_{\text{tgt}}, \varepsilon$
**Initialize:** $t \leftarrow 0, \lambda_j^{\{1\}} \leftarrow \lambda_{\text{tgt}}, \mu \leftarrow 0.9, \alpha \leftarrow \frac{1}{4}$
**Repeat:**
  $\theta^{(t+\frac{1}{2})} \leftarrow \operatorname{argmin}_\theta \mathcal{Q}(\theta; \theta^{(t)}, \lambda^{\{1\}})$
  $\Delta\theta^{(t)} \leftarrow \theta^{(t+\frac{1}{2})} - \theta^{(t)}$
  $\gamma_t \leftarrow \nabla\mathcal{L}\left(\theta^{(t)}\right)^\top \cdot \Delta\theta^{(t)} + \|\lambda^{\{1\}} \odot \left(\theta^{(t)} + \Delta\theta^{(t)}\right)\|_1 - \|\lambda^{\{1\}} \odot \theta^{(t)}\|_1$
  $\eta_t \leftarrow 1, q \leftarrow 0$
  **Repeat:**
    $\eta_t \leftarrow \mu^q$
    $q \leftarrow q + 1$
  **Until** $\mathcal{F}_{\lambda^{\{1\}}}\left(\theta^{(t)} + \eta_t \Delta\theta^{(t)}\right) \leq \mathcal{F}_{\lambda^{\{1\}}}\left(\theta^{(t)}\right) + \alpha\eta_t\gamma_t$
  $\theta^{(t+1)} \leftarrow \theta^{(t)} + \eta_t \Delta\theta^{(t)}$
  $t \leftarrow t + 1$
**Until** $\omega_{\lambda^{\{1\}}}(\theta^{(t)}) \leq \varepsilon$
**Return:** $\theta^{(t)}$.

---

## 3 Computational and Statistical Theories

Before we present our theoretical analysis, we first introduce some preliminaries, including important definitions and assumptions. We define the largest and smallest $s$-sparse eigenvalues of the Hessian matrix as follows.

**Definition 3.1.** *Given any positive integer $s$, we define the largest and smallest s-**sparse eigenvalues** of $\nabla^2\mathcal{L}(\theta)$ as*

$$\rho_s^+ = \sup_{\|v\|_0 \leq s} \frac{v^\top \nabla^2\mathcal{L}(\theta)v}{v^\top v} \quad \text{and} \quad \rho_s^- = \inf_{\|v\|_0 \leq s} \frac{v^\top \nabla^2\mathcal{L}(\theta)v}{v^\top v}.$$

**Algorithm 2** DC Proximal Newton Algorithm
---
**Input:** $\widehat{\theta}^{\{0\}}, \lambda_{\text{tgt}}, \beta, \varepsilon$
**Warm Initialization:** $\widehat{\theta}^{\{1\}} \leftarrow \text{ProxNewton}(\widehat{\theta}^{\{0\}}, \lambda_{\text{tgt}}, \varepsilon), K \leftarrow 1$
**Repeat:**
$\quad \lambda_j^{\{K+1\}} \leftarrow \begin{cases} 0, & \text{if } |\widehat{\theta}_j^{\{K\}}| > \beta \lambda_{\text{tgt}} \\ \lambda_{\text{tgt}}, & \text{if } |\widehat{\theta}_j^{\{K\}}| \leq \beta \lambda_{\text{tgt}} \end{cases}$
$\quad t \leftarrow 0, \theta^{(0)} = \widehat{\theta}^{\{K\}}$
$\quad$ **Repeat:**
$\quad\quad \theta^{(t+1)} \leftarrow \text{argmin}_\theta \mathcal{Q}(\theta; \theta^{(t)}, \lambda^{\{K+1\}})$
$\quad\quad t \leftarrow t + 1$
$\quad$ **Until** $\omega_{\lambda^{\{K+1\}}}(\theta^{(t)}) \leq \varepsilon$
$\quad \widehat{\theta}^{\{K+1\}} \leftarrow \theta^{(t)}$
$\quad K \leftarrow K + 1$
**Until** Convergence
**Return:** $\widehat{\theta}^{\{K\}}$.
---

Moreover, we define $\kappa_s = \rho_s^+ / \rho_s^-$ as the *s*-sparse condition number.

The sparse eigenvalue (SE) properties are widely studied in high dimensional sparse modeling problems, and are closely related to restricted strong convexity/smoothness properties and restricted eigenvalue properties (Zhou, 2009; van de Geer and Bühlmann, 2009; Raskutti et al., 2010; Negahban et al., 2012). For notational convenience, given a parameter $\theta \in \mathbb{R}^d$ and a real constant $R > 0$, we define a neighborhood of $\theta$ with radius $R$ as

$$\mathcal{B}(\theta, R) = \left\{\phi \in \mathbb{R}^d \mid \|\phi - \theta\|_2 \leq R\right\}.$$

Our first assumption is for the sparse eigenvalues of the Hessian matrix over a sparse domain.

**Assumption 1.** *Given $\theta \in \mathcal{B}(\theta^*, R)$ for a generic constant R, there exists a generic constant $C_0$ such that $\nabla^2 \mathcal{L}(\theta)$ satisfies the SE properties with parameters $\rho_{s^*+2\widetilde{s}}^-$ and $\rho_{s^*+2\widetilde{s}}^+$ satisfying*

$$0 < \rho_{s^*+2\widetilde{s}}^- < \rho_{s^*+2\widetilde{s}}^+ < +\infty \quad \text{with} \quad \widetilde{s} \geq C_0 \kappa_{s^*+2\widetilde{s}}^2 s^* \quad \text{and} \quad \kappa_{s^*+2\widetilde{s}} = \rho_{s^*+2\widetilde{s}}^+ / \rho_{s^*+2\widetilde{s}}^-.$$

Assumption 1 requires that $\nabla^2 \mathcal{L}(\theta)$ has finite largest and positive smallest sparse eigenvalues, given that $\theta$ is sufficiently sparse and close to $\theta^*$. Similar conditions are widely applied in the analyses of efficient algorithms for solving high dimensional learning problems, such as proximal gradient and coordinate gradient descent algorithms (Xiao and Zhang, 2013; Wang et al., 2014; Zhao et al., 2014; Li et al., 2016a,b). A direct consequence of Assumption 1 is the *restricted strong convexity/smoothness* of $\mathcal{L}(\theta)$ (RSC/RSS, Bühlmann and Van De Geer (2011)). Given any $\theta, \theta' \in \mathbb{R}^d$, the RSC/RSS parameter can be defined as

$$\delta(\theta', \theta) = \mathcal{L}(\theta') - \mathcal{L}(\theta) - \nabla \mathcal{L}(\theta)^\top (\theta' - \theta).$$

For notational simplicity, we define

$$\mathcal{S} = \{j \mid \theta_j^* \neq 0\} \text{ and } \mathcal{S}_\perp = \{j \mid \theta_j^* = 0\}.$$

The following proposition connects the SE properties to the RSC/RSS property.

**Proposition 3.2.** *Given $\theta, \theta' \in \mathcal{B}(\theta^*, R)$ with $\|\theta_{\mathcal{S}_\perp}\|_0 \leq \widetilde{s}$ and $\|\theta'_{\mathcal{S}_\perp}\|_0 \leq \widetilde{s}$, $\mathcal{L}(\theta)$ satisfies*

$$\frac{1}{2}\rho^-_{s^*+2\widetilde{s}}\|\theta' - \theta\|_2^2 \leq \delta(\theta', \theta) \leq \frac{1}{2}\rho^+_{s^*+2\widetilde{s}}\|\theta' - \theta\|_2^2.$$

The proof of Proposition 3.2 is provided in Bühlmann and Van De Geer (2011), and therefore is omitted. Proposition 3.2 implies that $\mathcal{L}(\theta)$ is essentially strongly convex, but only over a sparse domain (See Figure 2).

The second assumption requires $\nabla^2 \mathcal{L}(\theta)$ to be smooth over the sparse domain.

**Assumption 2** (Local Restricted Hessian Smoothness). *Recall that $\widetilde{s}$ is defined in Assumption 1. There exist generic constants $L_{s^*+2\widetilde{s}}$ and $R$ such that for any $\theta, \theta' \in \mathcal{B}(\theta^*, R)$ with $\|\theta_{\mathcal{S}_\perp}\|_0 \leq \widetilde{s}$ and $\|\theta'_{\mathcal{S}_\perp}\|_0 \leq \widetilde{s}$, we have*

$$\sup_{v \in \Omega, \|v\|=1} v^\top (\nabla^2 \mathcal{L}(\theta') - \nabla^2 \mathcal{L}(\theta)) v \leq L_{s^*+2\widetilde{s}} \|\theta - \theta'\|_2^2,$$

*where $\Omega = \{v \mid \text{supp}(v) \subseteq (\text{supp}(\theta) \cup \text{supp}(\theta'))\}$.*

Assumption 2 guarantees that $\nabla^2 \mathcal{L}(\theta)$ is Lipschitz continuous within a neighborhood of $\theta^*$ over a sparse domain. The local restricted Hessian smoothness is parallel to the local Hessian smoothness for analyzing the proximal Newton method in low dimensions (Lee et al., 2014), which is also close related to the self-concordance (Nemirovski, 2004) in the analysis of Newton method (Boyd and Vandenberghe, 2009).

In our analysis, we set the radius $R$ as

$$R = \frac{\rho^-_{s^*+2\widetilde{s}}}{2L_{s^*+2\widetilde{s}}}. \tag{3.1}$$

Note that $2R = \frac{\rho^-_{s^*+2\widetilde{s}}}{L_{s^*+2\widetilde{s}}}$ is the radius of the region centered at the unique sparse global minimizer of (2.2) for quadratic convergence of the proximal Newton algorithm, which will be further discussed later. This is parallel to the convergent radius in low dimensions (Lee et al., 2014), except that we restrict the parameters over the sparse domain.

The third assumption requires $\lambda_{\text{tgt}}$ to be chosen appropriately.

**Assumption 3.** *Given the true modeling parameter $\theta^*$, there exist generic constant $C_1$ such that*

$$\lambda_{\text{tgt}} = C_1 \sqrt{\frac{\log d}{n}} \geq 4\|\nabla \mathcal{L}(\theta^*)\|_\infty.$$

*Moreover, for large enough n, we have*

$$\sqrt{s^*} \lambda_{\text{tgt}} \leq C_2 R \rho^-_{s^*+2\widetilde{s}}.$$

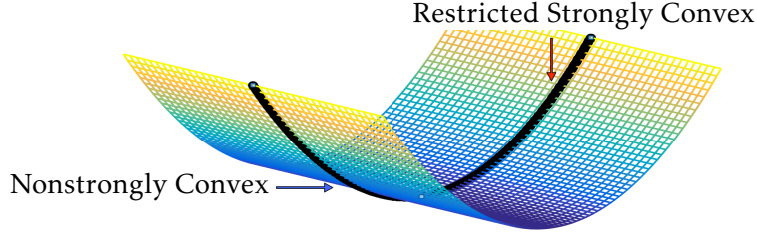

Figure 2: An illustrative two dimensional example of the restricted strong convexity. $\mathcal{L}(\theta)$ is not strongly convex. But if we restrict $\theta$ to be sparse (Black Curve), $\mathcal{L}(\theta)$ behaves like a strongly convex function.

Assumption 3 guarantees that the regularization is sufficiently large to eliminate irrelevant coordinates such that the obtained solution is sufficiently sparse (Bickel et al., 2009; Negahban et al., 2012). In addition, $\lambda_{\text{tgt}}$ can not be too large, which guarantees that the estimators are close enough to the true model parameter. The above assumptions are deterministic. We will verify these assumptions under GLM in the statistical analysis.

Our last assumption is on the predefined precision parameter $\varepsilon$ as follows.

**Assumption 4.** *For each stage of solving the convex relaxed subproblem* (2.2) *for all $K \geq 1$, we set*

$$\varepsilon = \frac{C_3}{\sqrt{n}} \leq \frac{\lambda_{\text{tgt}}}{8} \quad \text{for some generic small constant } C_3.$$

Assumption 4 guarantees that the output solution $\widehat{\theta}^{\{K\}}$ at each stage for all $K \geq 1$ has a sufficient precision, which is critical to our convergence analysis of multistage convex relaxation.

## 3.1 Computational Theory

We first characterize the convergence for the first stage of our proposed DC proximal Newton algorithm, i.e., the warm initialization for solving (2.3).

**Theorem 3.3** (Warm Initialization, $K = 1$). *Suppose that Assumptions 1 $\sim$ 4 hold with R defined in* (3.1). *After sufficiently many iterations $T < \infty$, the following results hold for all $t \geq T$:*

$$\|\theta^{(t)} - \theta^*\|_2 \leq R \quad \text{and} \quad \mathcal{F}_{\lambda^{\{1\}}}(\theta^{(t)}) \leq \mathcal{F}_{\lambda^{\{1\}}}(\theta^*) + \frac{15\lambda_{\text{tgt}}^2 s^*}{4\rho^-_{s^*+2\widetilde{s}}},$$

*which further guarantee*

$$\eta_t = 1, \ \|\theta^{(t)}_{\mathcal{S}_\perp}\|_0 \leq \widetilde{s}, \ \text{and} \ \|\theta^{(t+1)} - \overline{\theta}^{\{1\}}\|_2 \leq \frac{L_{s^*+2\widetilde{s}}}{2\rho^-_{s^*+2\widetilde{s}}} \|\theta^{(t)} - \overline{\theta}^{\{1\}}\|_2^2,$$

*where $\overline{\theta}^{\{1\}}$ is the unique sparse global minimizer of* (2.3) *satisfying $\|\overline{\theta}^{\{1\}}_{\mathcal{S}_\perp}\|_0 \leq \widetilde{s}$ and $\omega_{\lambda^{\{1\}}}(\overline{\theta}^{\{1\}}) = 0$. Moreover, we need at most*

$$T + \log\log\left(3\rho^+_{s^*+2\widetilde{s}}/\varepsilon\right)$$

iterations to terminate the proximal Newton algorithm for the warm initialization (2.3), where the output solution $\widehat{\theta}^{\{1\}}$ satisfies

$$\|\widehat{\theta}^{\{1\}}_{\mathcal{S}_\perp}\|_0 \leq \widetilde{s}, \ \omega_{\lambda^{\{1\}}}(\widehat{\theta}^{\{1\}}) \leq \varepsilon, \ \text{and} \ \|\widehat{\theta}^{\{1\}} - \theta^*\|_2 \leq \frac{18\lambda_{\text{tgt}}\sqrt{s^*}}{\rho^-_{s^*+2\widetilde{s}}}.$$

The proof of Theorem 3.3 is provided in Appendix 6.1. Theorem 3.3 implies:

(1) The objective value is sufficiently small after finite $T$ iterations of the proximal Newton algorithm, which further guarantees solutions to be sparse as well as good computational performance in all follow-up iterations.

(2) The solution enters the ball $\mathcal{B}(\theta^*, R)$ after finite $T$ iterations. Combined with the sparsity of the solution, it further guarantees that the solution enters the region of quadratic convergence. Thus the backtracking line search stops immediately and output $\eta_t = 1$ for all $t \geq T$.

(3) The total number of iterations is at most $\mathcal{O}(T + \log\log(1/\varepsilon))$ to achieve the approximate KKT condition $\omega_{\lambda^{\{1\}}}(\theta^{(t)}) \leq \varepsilon$, which serves as the stopping criterion of the warm initialization (2.3).

**Remark 3.4.** *To eliminate the notational ambiguity, we emphasis again the difference between $\overline{\theta}^{\{1\}}$ and $\widehat{\theta}^{\{1\}}$: $\overline{\theta}^{\{1\}}$ is the unique sparse global minimizer of (2.3) that satisfies the KKT condition, i.e., $\omega_{\lambda^{\{1\}}}(\overline{\theta}^{\{1\}}) = 0$; $\widehat{\theta}^{\{1\}}$ is the output solution of Algorithm 1 that satisfies the approximate KKT condition, i.e., $\omega_{\lambda^{\{1\}}}(\widehat{\theta}^{\{1\}}) \leq \varepsilon$ for some predefined $\varepsilon > 0$. Notations $\overline{\theta}^{\{K\}}$ and $\widehat{\theta}^{\{K\}}$ with the same interpretations above are also used for later stages $K \geq 2$.*

Given these good properties of the output solution $\widehat{\theta}^{\{1\}}$ obtained from the warm initialization, we can further show that our proposed DC proximal Newton algorithm for all follow-up stages (i.e., $K \geq 2$) achieves better computational performance than the first stage. This is characterized by the following theorem. For notational simplicity, we omit the iteration index $\{K\}$ for the intermediate updates within each stage for the multistage convex relaxation with $K \geq 2$.

**Theorem 3.5** (Stage $K$, $K \geq 2$). *Suppose Assumptions 1 $\sim$ 4 hold with R defined in (3.1). Then within each stage $K \geq 2$, for all iterations $t = 1, 2, ...,$ we have*

$$\|\theta^{(t)}_{\mathcal{S}_\perp}\|_0 \leq \widetilde{s} \quad \text{and} \quad \|\theta^{(t)} - \theta^*\|_2 \leq R,$$

*which further guarantee*

$$\eta_t = 1, \ \|\theta^{(t+1)} - \overline{\theta}^{\{K\}}\|_2 \leq \frac{L_{s^*+2\widetilde{s}}}{2\rho^-_{s^*+2\widetilde{s}}}\|\theta^{(t)} - \overline{\theta}^{\{K\}}\|_2^2, \ \text{and} \ \mathcal{F}_{\lambda^{\{K\}}}(\theta^{(t+1)}) < \mathcal{F}_{\lambda^{\{K\}}}(\theta^{(t)}),$$

*where $\overline{\theta}^{\{K\}}$ is the unique sparse global minimizer of (2.2) at the K-th stage satisfying $\|\overline{\theta}^{\{K\}}_{\mathcal{S}_\perp}\|_0 \leq \widetilde{s}$ and $\omega_{\lambda^{\{K\}}}(\overline{\theta}^{\{K\}}) = 0$. Moreover, we need at most*

$$\log\log\left(3\rho^+_{s^*+2\widetilde{s}}/\varepsilon\right).$$

*iterations to terminate the proximal Newton algorithm for the K-th stage of convex relaxation* (2.2), *where the output solution* $\widehat{\theta}^{\{K\}}$ *satisfies* $\|\widehat{\theta}^{\{K\}}_{\mathcal{S}_\perp}\|_0 \leq \widetilde{s}$, $\omega_{\lambda^{\{K\}}}(\widehat{\theta}^{\{K\}}) \leq \varepsilon$, *and*

$$\|\widehat{\theta}^{\{K\}} - \theta^*\|_2 \leq C_2 \left( \|\nabla \mathcal{L}(\theta^*)_{\mathcal{S}}\|_2 + \lambda_{\text{tgt}} \sqrt{\sum_{j \in \mathcal{S}} \mathbb{1}(|\theta^*_j| \leq \beta \lambda_{\text{tgt}})} + \varepsilon \sqrt{s^*} \right) + C_3 0.7^{K-1} \|\widehat{\theta}^{\{1\}} - \theta^*\|_2,$$

*for some generic constants* $C_2$ *and* $C_3$.

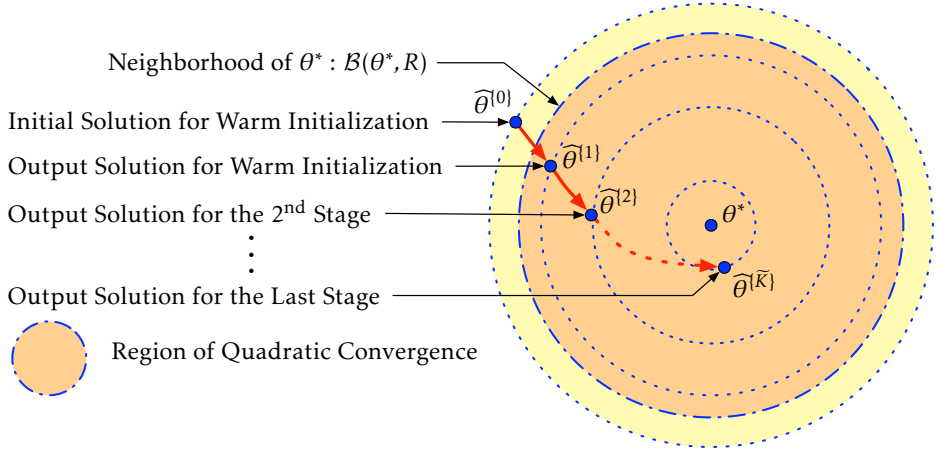

Figure 3: A geometric interpretation of local quadratic convergence: the warm initialization enters the region of quadratic convergence (orange region) after finite iterations and all follow-up stages remain in the region of quadratic convergence. The final estimator $\widehat{\theta}^{\{\widetilde{K}\}}$ has a better estimation error than the estimator $\widehat{\theta}^{\{1\}}$ obtained from the convex warm initialization.

The proof of Theorem 3.5 is provided in Appendix 6.2. A geometric interpretation for the computational theory of local quadratic convergence for our proposed algorithm is provided in Figure 3. Within each stage of the convex relaxation (2.2) for all $K \geq 2$, Theorem 3.5 implies:

(1) The algorithm maintains a sparse solution throughout all iterations $t \geq 1$. The sparsity further guarantees that the SE properties and local restricted Hessian smoothness hold, which are necessary conditions for the fast convergence of the proximal Newton algorithm.

(2) The solution is maintained in the region $\mathcal{B}(\theta^*, R)$ for all $t \geq 1$. Combined with the sparsity of the solution, we have that the solution enters the region of quadratic convergence. This guarantees that we only need to set the step size $\eta_t = 1$ and the objective value is monotone decreasing without the sophisticated backtracking line search procedure. Thus, the proximal Newton algorithm enjoys the same fast convergence as in low dimensional optimization problems (Lee et al., 2014).

(3) With the quadratic convergence rate, the number of iterations is at most $\mathcal{O}(\log \log(1/\varepsilon))$ to attain the approximate KKT condition $\omega_{\lambda^{\{K\}}}(\theta^{(t)}) \leq \varepsilon$, which is the stopping criteria at each stage.

## 3.2 Statistical Theory

Recall that our computational theory relies on deterministic assumptions (Assumptions 1 ∼ 3). However, these assumptions involve data, which are sampled from certain statistical distribution. Therefore, we need to verify that these assumptions hold with high probability under mild data generation process (e.g., GLM) in high dimensions in the following lemma.

**Lemma 3.6** (GLM). *Suppose that $x_i$'s are i.i.d. sampled from a zero-mean distribution with covariance matrix $\mathrm{Cov}(x_i) = \Sigma$ such that $\infty > c_{\max} \geq \Lambda_{\max}(\Sigma) \geq \Lambda_{\min}(\Sigma) \geq c_{\min} > 0$, and for any $v \in \mathbb{R}^d$, $v^\top x_i$ is sub-Gaussian with parameter at most $a\|v\|_2^2$, where $c_{\max}, c_{\min}$, and $a$ are generic constants. Moreover, for some constant $M_\psi > 0$, at least one of the following two conditions holds:*

(1) *The Hessian of the cumulant function $\psi$ is uniformly bounded: $\|\psi''\|_\infty \leq M_\psi$, or*

(2) *The covariates are bounded $\|x_i\|_\infty \leq 1$, and*

$$\mathbb{E}[\max_{|u|\leq 1}[\psi''(x^\top \theta^*) + u]^p] \leq M_\psi \quad \text{for some} \quad p > 2.$$

*Then Assumptions 1 ∼ 3 hold with high probability.*

The proof of Lemma 3.6 is provided in Appendix D. Given that these assumptions hold with high probability, the computational theory holds, i.e., the proximal Newton algorithm attains quadratic rate convergence within each stage of convex relaxation with high probability. We then further establish the statistical rate of convergence for the obtained estimator in parameter estimation.

**Theorem 3.7.** *Suppose the observations are generated from GLM satisfying the conditions in Lemma 3.6 for large enough $n$ such that $n \geq C_4 s^* \log d$ and $\beta = C_5/c_{\min}$ is a constant defined in (2.1) for generic constants $C_4$ and $C_5$, then with high probability, the output solution $\widehat{\theta}^{\{K\}}$ satisfies*

$$\|\widehat{\theta}^{\{K\}} - \theta^*\|_2 \leq C_6\left(\sqrt{\frac{s^*}{n}} + \sqrt{\frac{s' \log d}{n}}\right) + C_7 0.7^K\left(\sqrt{\frac{s^* \log d}{n}}\right)$$

*for generic constants $C_6$ and $C_7$, where $s' = \sum_{j \in \mathcal{S}} \mathbb{1}(|\theta_j^*| \leq \beta \lambda_{\mathrm{tgt}})$.*

Theorem 3.7 is a direct result combining Theorem 3.5 and the analyses in Zhang (2010b). As can be seen, $s'$ is essentially the number of non-zero $\theta_j$'s with smaller magnitudes than $\beta \lambda_{\mathrm{tgt}}$, which are often considered as "weak" signals. Theorem 3.7 essentially implies that by exploiting the multi-stage convex relaxation framework, our DC proximal Newton algorithm gradually reduces the estimation bias for "strong" signals, and eventually obtains an estimator with better statistical properties than the $\ell_1$-regularized estimator. Specifically, let $\widetilde{K}$ be the smallest integer such that after $\widetilde{K}$ stages of convex relaxation we have

$$C_7 0.7^{\widetilde{K}}\left(\sqrt{\frac{s^* \log d}{n}}\right) \leq C_6 \max\left\{\sqrt{\frac{s^*}{n}}, \sqrt{\frac{s' \log d}{n}}\right\},$$

which is equivalent to requiring $\widetilde{K} = \mathcal{O}(\log\log d)$. This implies the total number of the proximal Newton updates is at most

$$\mathcal{O}(T + \log\log(1/\varepsilon) \cdot (1 + \log\log d)).$$

In addition, the obtained estimator attains the optimal statistical properties in parameter estimation:

$$\|\widehat{\theta}^{\{\widetilde{K}\}} - \theta^*\|_2 \leq \mathcal{O}_P\left(\sqrt{\frac{s^*}{n}} + \sqrt{\frac{s'\log d}{n}}\right) \quad \text{v.s.} \quad \|\widehat{\theta}^{\{1\}} - \theta^*\|_2 \leq \mathcal{O}_P\left(\sqrt{\frac{s^*\log d}{n}}\right). \tag{3.2}$$

Recall that $\widehat{\theta}^{\{1\}}$ is obtained by the warm initialization (2.3). As illustrated in Figure 3, this implies the statistical rate in (3.2) for $\|\widehat{\theta}^{\{\widetilde{K}\}} - \theta^*\|_2$ obtained from the multistage convex relaxation for the nonconvex regularized problem (1.1) is a significant improvement over $\|\widehat{\theta}^{\{1\}} - \theta^*\|_2$ obtained from the convex problem (2.3). Especially when $s'$ is small, i.e., most of non-zero $\theta_j$'s are strong signals, our result approaches the oracle bound[3] $\mathcal{O}_P(\sqrt{s^*/n})$ (Fan and Li, 2001) as illustrated in Figure 4.

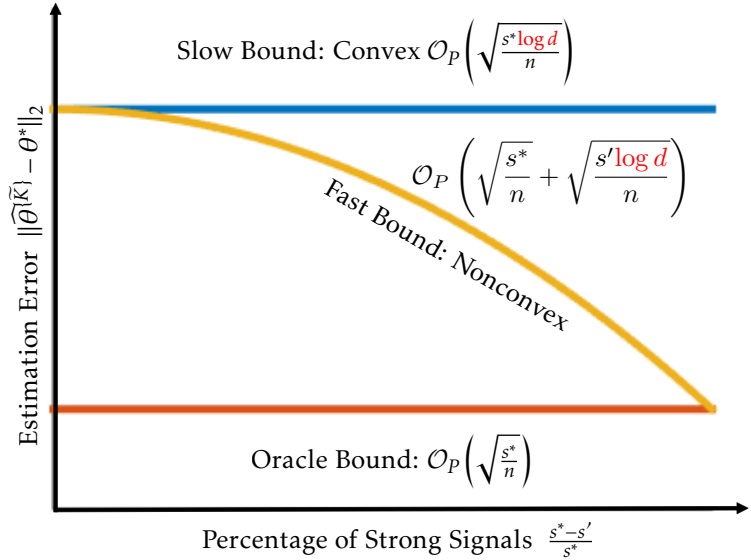

Figure 4: An illustration of the statistical rates of convergence in parameter estimation. Our obtained estimator has an error bound between the oracle bound and the slow bound from the convex problem in general. When the percentage of strong signals increases, i.e., $s'$ decreases, then our result approaches the oracle bound.

## 4 Experiments

We compare our DC Proximal Newton algorithm (DC+PN) with two competing algorithms for solving nonconvex regularized sparse logistic regression problems. They are accelerated proximal gradient algorithm (APG) implemented in the SPArse Modeling Software (SPAMS, coded in C++, Mairal et al. (2014)), and accelerated coordinate descent (ACD) algorithm implemented in R package gcdnet (coded in Fortran, Yang and Zou (2013)). We further optimize the active set strategy in gcdnet to boost its computational performance. To integrate these two algorithms with the multistage convex relaxation framework, we revise their source code.

---
[3]The oracle bound assumes that we know which variables are relevant in advance. It is not a realistic bound, but only for comparison purpose.

To further boost the computational efficiency at each stage of the convex relaxation, , we apply the pathwise optimization for all algorithms (Friedman et al., 2010; Zhao et al., 2014). Specifically, we use a geometrically decreasing sequence of regularization parameters $\{\lambda_{[m]} = \alpha^m \lambda_{[0]}\}_{m=1}^M$, where $\alpha \in (0,1)$ is a shrinkage ratio, and $\lambda_{\text{tgt}} = \lambda_{[M]}$. For each $\lambda_{[m]}$, we apply the corresponding algorithm (DC+PN, DC+APG, and DC+ACD) to solve the nonconvex regularized problem (1.1). The value of $\lambda_{[0]}$ is chosen to be the smallest value such that the corresponding solution is zero. Moreover, we initialize the solution for a new regularization parameter $\lambda_{[m+1]}$ using the output solution obtained with $\lambda_{[m]}$. Such a pathwise optimization scheme has achieved tremendous success in practice (Friedman et al., 2010; Zhao et al., 2012; Li et al., 2015), and we refer Zhao et al. (2014) for more involved theoretical analysis.

All three algorithms are compared in wall clock time and objective values with $\lambda_{\text{tgt}} \approx \frac{1}{4}\sqrt{\log d/n}$. Our DC Proximal Newton algorithm is implemented in C with double precisions, and called from R by a wrapper. Our comparison contains 3 datasets: "madelon" ($n = 2000, d = 500$, Guyon et al. (2005)), "gisette" ($n = 2000, d = 5000$, Guyon et al. (2005)), and three simulated datasets: "sim_1k", "sim_5k", and "sim_10k". For the simulated data sets, we choose $n = 1000$ and $d = 5000$, and generate each $x_i$ independently from a $d$-dimensional normal distribution $\mathcal{N}(0, \Sigma)$, where $\Sigma_{jk} = 0.5^{|j-k|}$ for all $j,k = 1,...,d$. We generate $y \sim \text{Bernoulli}(1/[1+\exp(-x_i^\top \theta^*)])$, where $\theta^*$ has all 0 entries except randomly selected 20 entries. The non-zero entries are independently sampled from U(0,1).

Table 1: Quantitive timing comparisons for on nonconvex-regularized sparse logistic regression. DC+PN denotes our proposed DC proximal Newton algorithm; ACD denotes the coordinate descent algorithm combined with the active set strategy; APG denotes the accelerated proximal gradient algorithm. The average values and standard errors (in parentheses) of timing performance (in seconds) are presented.

|        | madelon | gisette | sim_1k | sim_5k | sim_10k |
|--------|---------|---------|--------|--------|---------|
| DC+PN  | **1.51**(±0.01)s | **5.35**(±0.11)s | **1.07**(±0.02)s | **4.53**(±0.06)s | **8.82**(±0.04)s |
|        | obj value: 0.52 | obj value: 0.01 | obj value: 0.01 | obj value: 0.01 | obj value: 0.01 |
| DC+ACD | 5.83(±0.03)s | 18.92(±2.25)s | 9.46(±0.09) s | 16.20(±0.24) s | 19.1(±0.56) s |
|        | obj value: 0.52 | obj value: 0.01 | obj value: 0.01 | obj value: 0.01 | obj value: 0.01 |
| DC+APG | 1.60(±0.03)s | 207(±2.25)s | 17.8(±1.23) s | 111(±1.28) s | 222(±5.79) s |
|        | obj value: 0.52 | obj value: 0.01 | obj value: 0.01 | obj value: 0.01 | obj value: 0.01 |

The experiments are performed on a personal computer with 2.6GHz Intel Core i7 and 16GB RAM. For each algorithm and dataset, we repeat the algorithm 10 times and we report the aver-age values and standard errors of the wall clock time in Table 1. The stopping criteria for each algorithms are tuned such that they attains similar optimization errors. As can be seen in Table 1, our DC Proximal Newton algorithm significantly outperforms the competing algorithms in terms of the timing performance.

We then illustrate the quadratic convergence of our DC proximal Newton algorithm within each stage of convex relaxation using the "sim" datasets. Specifically, we provide the plots of gap towards the optimal objective of the $K$-th stage, i.e., $\log(\mathcal{F}_{\lambda^{\{K\}}}(\theta^{(t)}) - \mathcal{F}_{\lambda^{\{K\}}}(\overline{\theta}^{\{K\}}))$, for $K = 1, 2, 3, 4$ in a single simulation in Figure 5. We see that our DC proximal Newton algorithm achieves quadratic convergence, which is consistent with our theory.

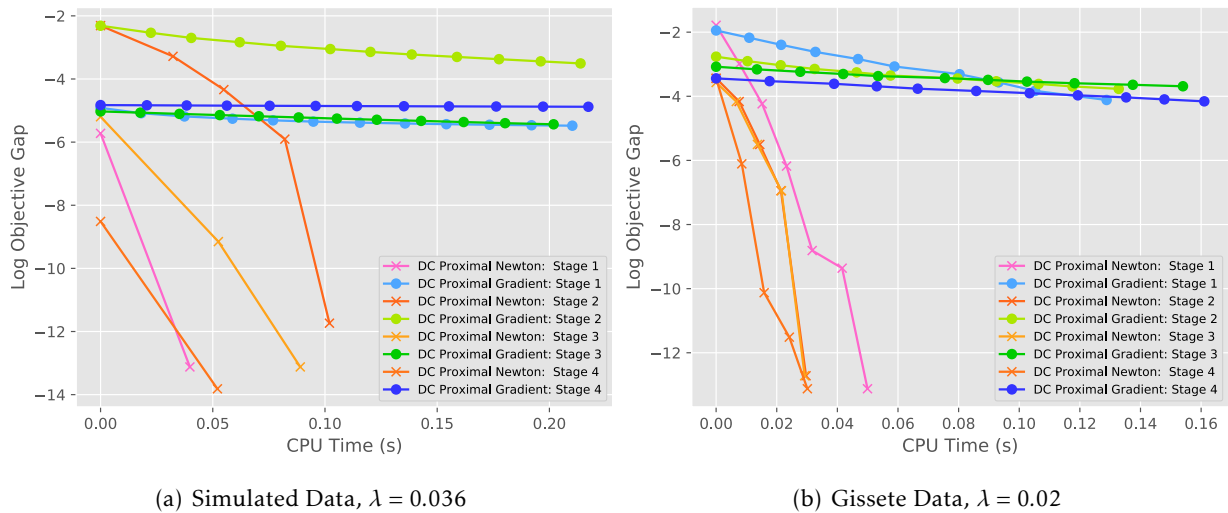

(a) Simulated Data, $\lambda = 0.036$

(b) Gissete Data, $\lambda = 0.02$

Figure 5: Timing comparisons in wall clock time. Our DC proximal Newton algorithm demonstrates superior quadratic convergence (consistent with our theory), and significantly outperforms the DC proximal gradient algorithm.

## 5 Discussions and Future Work

We first provide detailed discussions on the superior performance of our DC proximal Newton in our experiment, and then discuss potential variants – DC proximal sub-sampled Newton or Quasi-Newton algorithm.

### 5.1 Drawbacks of first order algorithms

There exist two major drawbacks of existing multi-stage convex relaxation based first order algorithms:

**(1)** The first order algorithms have significant computational overhead in each iteration, e.g., for GLM, computing gradients requires frequently evaluating the cumulant function and its derivatives. This often involves extensive non-arithmetic operations such as `log` and `exp` functions, which naturally appear in the cumulant function and its derivates and are computationally expensive. To the best of our knowledge, even if we use some efficient numerical methods for calculating `exp` in Schraudolph (1999); Malossi et al. (2015), the computation still needs at least 10 – 30 times more CPU cycles than basic arithmetic operations, e.g., multiplications. Our proposed DC Proximal Newton algorithm cannot avoid calculating the cumulant function and its derivatives, when computing quadratic approximations. The computation, however, is much less intense, since the convergence is quadratic.

(2) The first order algorithms are computationally expensive with the step size selection. Although for certain GLM, e.g., sparse logistic regression, we can choose the step size parameter as

$$\eta \approx \Lambda_{\max}^{-1}\left(\frac{1}{n}\sum_{i=1}^{n} x_i x_i^\top\right).$$

However, such a step size often leads to very poor performance. In contrast, as our theoretical analysis and experiments suggest, the proposed DC proximal Newton algorithm needs very few line search steps, which saves much computational efforts.

Some recent papers on proximal Newton or inexact proximal Newton also demonstrate local quadratic convergence guarantees, such as Yen et al. (2014); Yue et al. (2016). However, their conditions are much more stringent than the SE properties in terms of the dependence on the problem dimensions. Specifically, their quadratic convergence can only be guaranteed on a much smaller ball/neighborhood. For example, the constant nullspace strong convexity in Yen et al. (2014), which plays the same role as the smallest sparse eigenvalue $\rho_{s^*+2\widetilde{s}}^-$ in our analysis, is as small as $1/d$. Thus, they can only guarantee the quadratic convergence in a region with radius $\mathcal{O}(1/d)$, which is very small in high dimensions. However, in our analysis, $\rho_{s^*+2\widetilde{s}}^-$ can be a constant, which is (almost) independent of $d$ (Bühlmann and Van De Geer, 2011) and much larger than $\mathcal{O}(1/d)$. A similar issue that the quadratic region is too small exists in Yue et al. (2016) as well.

## 5.2 Extension to sub-sampled or Quasi-Newton algorithms

Our methodology can be further extended to proximal sub-sampled Newton or Quasi-Newton algorithms using either BFGS-type or subsampled Hessian matrices. Taking the Proximal sub-sampled Newton algorithm as an example, we replace the Hessian matrix with an approximate Hessian matrix in each proximal Newton iteration. Suppose that at the $t$-th iteration of the $K$-th stage, we randomly select a mini-batch $\mathcal{X}^{(t)} \subset \{1,...,n\}$ of $m$ samples from the data with equal probability (i.e., $|\mathcal{X}^{(t)}| = m$). We then consider an alternative quadratic approximation

$$\widehat{\mathcal{Q}}(\theta;\theta^{(t)},\lambda^{(K)},\mathcal{X}^{(t)}) = \mathcal{L}(\theta^{(t)}) + (\theta - \theta^{(t)})^\top \nabla \mathcal{L}(\theta^{(t)}) + \frac{1}{2}\|\theta - \theta^{(t)}\|^2_{\widehat{H}(\theta^{(t)},\mathcal{X}^{(t)})} + \|\lambda^{\{K\}} \odot \theta\|_1, \quad (5.1)$$

where $\widehat{H}(\theta^{(t)}, \mathcal{X}^{(t)})$ is the subsampled Hessian matrix

$$\widehat{H}(\theta^{(t)}, \mathcal{X}^{(t)}) = \frac{1}{m} \sum_{i \in \mathcal{X}^{(t)}} \nabla^2 \ell_i(\theta^{(t)}).$$

By exploiting the additive nature of $\mathcal{L}(\theta)$, we can further rewrite (5.1) as

$$\widehat{\mathcal{Q}}(\theta; \theta^{(t)}, \lambda^{(K)}, \mathcal{X}^{(t)}) = \frac{1}{m} \sum_{i \in \mathcal{X}^{(t)}} w_i (x_i^\top \theta)^2 + g^\top \theta + \|\lambda^{\{K\}} \odot \theta\|_1 + \text{constant}, \tag{5.2}$$

where $g \in \mathbb{R}^d$ and $w_i \in \mathbb{R}$ for all $i \in \mathcal{X}^{(t)}$ are some easy to compute constants depending on $\theta^{(t)}$, $\ell_i(\theta^{(t)})$'s, $x_i$'s, and $y_i$'s. Similar to (2.5), (5.2) only requires $O(md)$ memory usage and can be efficiently solved by coordinate descent algorithms in conjunction with the active set strategy, soft thresholding, and residual update. See more details in Appendix A. Note that the line search procedure is needed for the proximal sub-sampled Newton algorithm throughout all iterations and stages.

The sub-sampled Hessian matrices preserve the spectral behaviors when the batch size $m$ is large enough (e.g. $m = \Omega(s^* \log d)$). Thus, restricted strong convexity, smoothness, and Hessian smoothness hold, and similar theoretical results are expected to hold. A major difference is that we get slower convergence (e.g. superlinear or linear depending on the batch size $m$) instead of quadratic convergence. This is a fundamental tradeoff between Proximal Newton and proximal sub-sampled Newton (or Quasi-Newton) algorithm in both low and high dimensions. We will leave this for further investigation.

# 6 Proofs of Main Results

We provide proof sketches for the main results of Theorem 3.3 and 3.5 in this section.

## 6.1 Proof of Theorem 3.3

We provide a few important intermediate results. The first result characterizes the sparsity of the solution and an upper bound of the objective after sufficiently many iterations as follows. The proof is provided in Appendix C.1.

**Lemma 6.1.** *Suppose that Assumptions 1 $\sim$ 4 hold. After sufficiently many iterations $T < \infty$, the following results hold for all $t \geq T$:*

$$\|\theta^{(t)}_{\mathcal{S}_\perp}\|_0 \leq \widetilde{s} \quad \text{and} \quad \mathcal{F}_{\lambda^{\{1\}}}(\theta^{(t)}) \leq \mathcal{F}_{\lambda^{\{1\}}}(\theta^*) + \frac{15 \lambda_{\text{tgt}}^2 s^*}{4 \rho_{s^* + 2\widetilde{s}}^-}.$$

We then demonstrate the parameter estimation and quadratic convergence conditioning on the sparse solution and bounded objective. The proof is provided in Appendix C.2.

**Lemma 6.2.** *Suppose that Assumptions 1 ~ 4 hold. If $\|\theta^{(t)}_{\mathcal{S}_\perp}\|_0 \leq \widetilde{s}$, and $\mathcal{F}_{\lambda^{\{1\}}}(\theta^{(t)}) \leq \mathcal{F}_{\lambda^{\{1\}}}(\theta^*) + \frac{15\lambda^2_{\text{tgt}}s^*}{4\rho^-_{s^*+2\widetilde{s}}}$, we have*

$$\|\theta^{(t)} - \theta^*\|_2 \leq \frac{18\lambda_{\text{tgt}}\sqrt{s^*}}{\rho^-_{s^*+2\widetilde{s}}} \text{ and } \|\theta^{(t+1)} - \overline{\theta}^{\{1\}}\|_2 \leq \frac{L_{s^*+2\widetilde{s}}}{2\rho^-_{s^*+2\widetilde{s}}}\|\theta^{(t)} - \overline{\theta}^{\{1\}}\|^2_2$$

Moreover, we characterize the sufficient number of iterations for the proximal Newton updates to achieve the approximate KKT condition. The proof is provided in Appendix C.3.

**Lemma 6.3.** *Suppose that Assumptions 1 ~ 4 hold. If $\|\theta^{(T)}_{\mathcal{S}_\perp}\|_0 \leq \widetilde{s}$, and $\mathcal{F}_{\lambda^{\{1\}}}(\theta^{(T)}) \leq \mathcal{F}_{\lambda^{\{1\}}}(\theta^*) + \frac{15\lambda^2_{\text{tgt}}s^*}{4\rho^-_{s^*+2\widetilde{s}}}$ at some iteration T, we need at most*

$$T_1 \leq \log\log\left(\frac{3\rho^+_{s^*+2\widetilde{s}}}{\varepsilon}\right)$$

*extra iterations of the proximal Newton updates such that $\omega_{\lambda^{\{1\}}}(\theta^{(T+T_1)}) \leq \frac{\lambda_{\text{tgt}}}{8}$.*

Combining Lemma 6.1 ~ 6.3, we have desired results in Theorem 3.3.

## 6.2 Proof of Theorem 3.5

We present a few important intermediate results that are key components of our main proof. The first result shows that in a neighborhood of the true model parameter $\theta^*$, the sparsity of the solution is preserved when we use a sparse initialization. The proof is provided in Appendix B.1.

**Lemma 6.4** (Sparsity Preserving Lemma). *Suppose that Assumptions 1 and 2 hold with $\varepsilon \leq \frac{\lambda_{\text{tgt}}}{8}$. Given $\theta^{(t)} \in \mathcal{B}(\theta^*, R)$ and $\|\theta^{(t)}_{\mathcal{S}_\perp}\|_0 \leq \widetilde{s}$, there exists a generic constant $C_1$ such that*

$$\|\theta^{(t+1)}_{\mathcal{S}_\perp}\|_0 \leq \widetilde{s} \text{ and } \|\theta^{(t+1)} - \theta^*\|_2 \leq \frac{C_1\lambda_{\text{tgt}}\sqrt{s^*}}{\rho^-_{s^*+2\widetilde{s}}}.$$

We then show that every step of proximal Newton updates within each stage has a quadratic convergence rate to a local minimizer, if we start with a sparse solution in the refined region. The proof is provided in Appendix B.2.

**Lemma 6.5.** *Suppose that Assumptions 1 ~ 4 hold. If $\theta^{(t)} \in \mathcal{B}(\theta^*, R)$ and $\|\theta^{(t)}_{\mathcal{S}_\perp}\|_0 \leq \widetilde{s}$, then for each stage $K \geq 2$, we have*

$$\|\theta^{(t+1)} - \overline{\theta}^{\{K\}}\|_2 \leq \frac{L_{s^*+2\widetilde{s}}}{2\rho^-_{s^*+2\widetilde{s}}}\|\theta^{(t)} - \overline{\theta}^{\{K\}}\|^2_2.$$

In the following, we need to use the property that the iterates $\theta^{(t)} \in \mathcal{B}(\overline{\theta}^{\{K\}}, 2R)$ instead of $\theta^{(t)} \in \mathcal{B}(\theta^*, R)$ for convergence analysis of the proximal Newton method. This property holds since we have $\theta^{(t)} \in \mathcal{B}(\theta^*, R)$ and $\overline{\theta}^{\{K\}} \in \mathcal{B}(\theta^*, R)$ simultaneously. Thus $\theta^{(t)} \in \mathcal{B}\left(\overline{\theta}^{\{K\}}, 2R\right)$, where $2R = \frac{\rho^-_{s^*+2\widetilde{s}}}{L_{s^*+2\widetilde{s}}}$ is the radius for quadratic convergence region of the proximal Newton algorithm.

The following lemma demonstrates that the step size parameter is simply 1 if the the sparse solution is in the refined region. The proof is provided in Appendix B.3.

**Lemma 6.6.** *Suppose that Assumptions 1 ∼ 4 hold. If $\theta^{(t)} \in \mathcal{B}(\overline{\theta}^{\{K\}}, 2R)$ and $\|\theta^{(t)}_{\mathcal{S}_\perp}\|_0 \leq \widetilde{s}$ at each stage $K \geq 2$ with $\frac{1}{4} \leq \alpha < \frac{1}{2}$, then $\eta_t = 1$. Further, we have*

$$\mathcal{F}_{\lambda^{\{K\}}}(\theta^{(t+1)}) \leq \mathcal{F}_{\lambda^{\{K\}}}(\theta^{(t)}) + \frac{1}{4}\gamma_t.$$

Moreover, we present a critical property of $\gamma_t$. The proof is provided in Appendix B.4.

**Lemma 6.7.** *Denote $\Delta\theta^{(t)} = \theta^{(t)} - \theta^{(t+1)}$ and*

$$\gamma_t = \nabla\mathcal{L}(\theta^{(t)})^\top \cdot \Delta\theta^{(t)} + \|\lambda^{\{K\}} \odot (\theta^{(t)} + \Delta\theta^{(t)})\|_1 - \|\lambda^{\{K\}} \odot (\theta^{(t)})\|_1.$$

*Then we have $\gamma_t \leq -\|\Delta\theta^{(t)}\|^2_{\nabla^2\mathcal{L}(\theta^{(t)})}$.*

In addition, we present the sufficient number of iterations for each convex relaxation stage to achieve the approximate KKT condition. The proof is provided in Appendix B.5.

**Lemma 6.8.** *Suppose that Assumptions 1 ∼ 4 hold. To achieve the approximate KKT condition $\omega_{\lambda^{\{K\}}}(\theta^{(t)}) \leq \varepsilon$ for any $\varepsilon > 0$ at each stage $K \geq 2$, the number of iteration for proximal Newton updates is at most*

$$\log\log\left(\frac{3\rho^+_{s^*+2\widetilde{s}}}{\varepsilon}\right).$$

We further present the contraction of the estimation error along consecutive stages, which is a direct result from oracle statistical rate in Fan et al. (2015).

**Lemma 6.9.** *Suppose that Assumptions 1 ∼ 4 hold. Then there exists a generic constant $c_1$ such that the output solutions for all $K \geq 2$ satisfy*

$$\|\widehat{\theta}^{\{K\}} - \theta^*\|_2 \leq c_1 \left(\|\nabla\mathcal{L}(\theta^*)_\mathcal{S}\|_2 + \lambda_{\text{tgt}}\sqrt{\sum_{j\in\mathcal{S}}\mathbb{1}(|\theta^*_j| \leq \beta\lambda)} + \varepsilon\sqrt{s^*}\right) + 0.7\|\widehat{\theta}^{\{K-1\}} - \theta^*\|_2.$$

Combining Lemma 6.4 ∼ Lemma 6.7, we have the quadratic convergence of the proximal Newton algorithm within each convex relaxation stage. The rest of the results in Theorem 3.5 hold by further combining Lemma 6.8 and recursively applying Lemma 6.9.

# References

Armijo, L. (1966). Minimization of functions having lipschitz continuous first partial derivatives. *Pacific Journal of Mathematics* **16** 1–3.

Beck, A. and Teboulle, M. (2009). Fast gradient-based algorithms for constrained total variation image denoising and deblurring problems. *IEEE Transactions on Image Processing* **18** 2419–2434.

Belloni, A., Chernozhukov, V. and Wang, L. (2011). Square-root lasso: pivotal recovery of sparse signals via conic programming. *Biometrika* **98** 791–806.

Bickel, P. J., Ritov, Y. and Tsybakov, A. B. (2009). Simultaneous analysis of Lasso and Dantzig selector. *The Annals of Statistics* **37** 1705–1732.

Boyd, S. and Vandenberghe, L. (2009). *Convex Optimization*. Cambridge University Press.

Bühlmann, P. and Van De Geer, S. (2011). *Statistics for high-dimensional data: methods, theory and applications*. Springer Science & Business Media.

Eloyan, A., Muschelli, J., Nebel, M. B., Liu, H., Han, F., Zhao, T., Barber, A. D., Joel, S., Pekar, J. J., Mostofsky, S. H. et al. (2012). Automated diagnoses of attention deficit hyperactive disorder using magnetic resonance imaging. *Frontiers in systems neuroscience* **6**.

Fan, J. and Li, R. (2001). Variable selection via nonconcave penalized likelihood and its oracle properties. *Journal of the American Statistical Association* **96** 1348–1360.

Fan, J., Liu, H., Sun, Q. and Zhang, T. (2015). TAC for sparse learning: Simultaneous control of algorithmic complexity and statistical error. *arXiv preprint arXiv:1507.01037* .

Friedman, J., Hastie, T. and Tibshirani, R. (2010). Regularization paths for generalized linear models via coordinate descent. *Journal of Statistical Software* **33** 1.

Guyon, I., Gunn, S., Ben-Hur, A. and Dror, G. (2005). Result analysis of the nips 2003 feature selection challenge. In *Advances in neural information processing systems*.

Lee, J. D., Sun, Y. and Saunders, M. A. (2014). Proximal newton-type methods for minimizing composite functions. *SIAM Journal on Optimization* **24** 1420–1443.

Li, X., Haupt, J., Arora, R., Liu, H., Hong, M. and Zhao, T. (2016a). A first order free lunch for sqrt-lasso. *arXiv preprint arXiv:1605.07950* .

Li, X., Zhao, T., Arora, R., Liu, H. and Haupt, J. (2016b). Stochastic variance reduced optimization for nonconvex sparse learning. In *International Conference on Machine Learning*.

Li, X., Zhao, T., Zhang, T. and Liu, H. (2015). The picasso package for nonconvex regularized m-estimation in high dimensions in R. *Technical Report* .

Liu, H., Wang, L. and Zhao, T. (2015). Calibrated multivariate regression with application to neural semantic basis discovery. *Journal of Machine Learning Research* **16** 1579–1606.

Loh, P.-L. and Wainwright, M. J. (2015). Regularized m-estimators with nonconvexity: Statistical and algorithmic theory for local optima. *Journal of Machine Learning Research* To appear.

Luo, Z.-Q. and Tseng, P. (1992). On the linear convergence of descent methods for convex essentially smooth minimization. *SIAM Journal on Control and Optimization* **30** 408–425.

Mairal, J., Bach, F., Ponce, J. et al. (2014). Sparse modeling for image and vision processing. *Foundations and Trends® in Computer Graphics and Vision* **8** 85–283.

Malossi, A. C. I., Ineichen, Y., Bekas, C. and Curioni, A. (2015). Fast exponential computation on simd architectures. *Proc. of HIPEAC-WAPCO, Amsterdam NL* .

McCullagh, P. (1984). Generalized linear models. *European Journal of Operational Research* **16** 285–292.

Neale, B. M., Kou, Y., Liu, L., Ma'Ayan, A., Samocha, K. E., Sabo, A., Lin, C.-F., Stevens, C., Wang, L.-S., Makarov, V. et al. (2012). Patterns and rates of exonic de novo mutations in autism spectrum disorders. *Nature* **485** 242–245.

Negahban, S. N., Ravikumar, P., Wainwright, M. J. and Yu, B. (2012). A unified framework for high-dimensional analysis of m-estimators with decomposable regularizers. *Statistical Science* **27** 538–557.

Nemirovski, A. (2004). Interior point polynomial time methods in convex programming. *Lecture notes* .

Nesterov, Y. (2013). Gradient methods for minimizing composite functions. *Mathematical Programming* **140** 125–161.

Ning, Y., Zhao, T. and Liu, H. (2014). A likelihood ratio framework for high dimensional semi-parametric regression. *arXiv preprint arXiv:1412.2295* .

Pfanzagl, J. (1994). *Parametric statistical theory*. Walter de Gruyter.

Raginsky, M., Willett, R. M., Harmany, Z. T. and Marcia, R. F. (2010). Compressed sensing performance bounds under poisson noise. *IEEE Transactions on Signal Processing* **58** 3990–4002.

Raskutti, G., Wainwright, M. J. and Yu, B. (2010). Restricted eigenvalue properties for correlated Gaussian designs. *Journal of Machine Learning Research* **11** 2241–2259.

Schraudolph, N. N. (1999). A fast, compact approximation of the exponential function. *Neural Computation* **11** 853–862.

Shalev-Shwartz, S. and Tewari, A. (2011). Stochastic methods for $\ell_1$-regularized loss minimization. *Journal of Machine Learning Research* **12** 1865–1892.

Shen, X., Pan, W. and Zhu, Y. (2012). Likelihood-based selection and sharp parameter estimation. *Journal of the American Statistical Association* **107** 223–232.

Tibshirani, R. (1996). Regression shrinkage and selection via the lasso. *Journal of the Royal Statistical Society. Series B (Methodological)* 267–288.

Tibshirani, R., Bien, J., Friedman, J., Hastie, T., Simon, N., Taylor, J. and Tibshirani, R. J. (2012). Strong rules for discarding predictors in Lasso-type problems. *Journal of the Royal Statistical Society: Series B (Statistical Methodology)* **74** 245–266.

van de Geer, S. A. (2008). High-dimensional generalized linear models and the Lasso. *The Annals of Statistics* **36** 614–645.

van de Geer, S. A. and Bühlmann, P. (2009). On the conditions used to prove oracle results for the Lasso. *Electronic Journal of Statistics* **3** 1360–1392.

Wang, Z., Liu, H. and Zhang, T. (2014). Optimal computational and statistical rates of convergence for sparse nonconvex learning problems. *The Annals of Statistics* **42** 2164–2201.

Xiao, L. and Zhang, T. (2013). A proximal-gradient homotopy method for the sparse least-squares problem. *SIAM Journal on Optimization* **23** 1062–1091.

Yang, Y. and Zou, H. (2013). An efficient algorithm for computing the hhsvm and its generalizations. *Journal of Computational and Graphical Statistics* **22** 396–415.

Yen, I. E.-H., Hsieh, C.-J., Ravikumar, P. K. and Dhillon, I. S. (2014). Constant nullspace strong convexity and fast convergence of proximal methods under high-dimensional settings. In *Advances in Neural Information Processing Systems*.

Yue, M.-C., Zhou, Z. and So, A. M.-C. (2016). Inexact regularized proximal newton method: provable convergence guarantees for non-smooth convex minimization without strong convexity. *arXiv preprint arXiv:1605.07522* .

Zhang, C.-H. (2010a). Nearly unbiased variable selection under minimax concave penalty. *The Annals of Statistics* **38** 894–942.

Zhang, T. (2010b). Analysis of multi-stage convex relaxation for sparse regularization. *Journal of Machine Learning Research* **11** 1081–1107.

Zhao, T., Liu, H., Roeder, K., Lafferty, J. and Wasserman, L. (2012). The huge package for high-dimensional undirected graph estimation in R. *Journal of Machine Learning Research* **13** 1059–1062.

Zhao, T., Liu, H. and Zhang, T. (2014). Pathwise coordinate optimization for sparse learning: Algorithm and theory. *arXiv preprint arXiv:1412.7477* .

Zhou, S. (2009). Restricted eigenvalue conditions on subgaussian random matrices. *arXiv preprint arXiv:0912.4045* .

## A  Coordinate Descent Algorithms with the Active Set Strategy

We first provide a brief derivation of the quadratic approximation (2.4) into a weighted least square problem. For notational convenience, we omit the indices $\{K\}$ and $(t)$ for a particular iter-ation of a stage. Recall that we want to minimize the following $\ell_1$-regularized quadratic problem

$$\widehat{\Delta\theta} = \underset{\Delta\theta}{\operatorname{argmin}} \ \Delta\theta^\top \nabla\mathcal{L}(\theta) + \frac{1}{2}\Delta\theta^\top \nabla^2\mathcal{L}(\theta)\Delta\theta + \|\lambda \odot (\theta + \Delta\theta)\|_1. \tag{A.1}$$

For GLM, we have

$$\mathcal{L}(\theta) = \frac{1}{n}\sum_{i=1}^{n}\psi(x_i^\top\theta) - y_i x_i^\top\theta,$$

where $\psi$ is the cumulant function. Then we can rewrite the quadratic function $\Delta\theta^\top\nabla\mathcal{L}(\theta) + \frac{1}{2}\Delta\theta^\top\nabla^2\mathcal{L}(\theta)\Delta\theta$ in subproblem (A.1) as a *weighted least squares* form (Friedman et al., 2010):

$$\frac{1}{2n}\sum_{i=1}^{n}\left(2\left(y_i - \psi'(x_i^\top\theta)\right)x_i^\top\Delta\theta + \psi''(x_i^\top\theta)(x_i^\top\Delta\theta)^2\right) = \frac{1}{2n}\sum_{i=1}^{n}w_i(z_i - x_i^\top\Delta\theta)^2 + \text{constant},$$

where $w_i = \psi''(x_i^\top\theta)$, $z_i = \frac{y_i - \psi'(x_i^\top\theta)}{\psi''(x_i^\top\theta)}$, and the constant term does not depend on $\Delta\theta$. This indicates that (A.1) is equivalent to a Lasso problem with reweighted least square loss function:

$$\widehat{\Delta\theta} = \underset{\Delta\theta}{\operatorname{argmin}} \ \frac{1}{2n}\sum_{i=1}^{n}w_i(z_i - x_i^\top\Delta\theta)^2 + \|\lambda \odot (\theta + \Delta\theta)\|_1. \tag{A.2}$$

By solving (A.2), we can avoid directly computing the $d \times d$ Hessian matrix $\nabla^2\mathcal{L}(\theta)$ in (A.1) and significantly reduce the memory usage when $d$ is large.

We then introduce an algorithm for solving (A.2) leveraging the idea of active set update. The active set update scheme is very efficient in practice (Friedman et al., 2010) with rigid theoretical justifications (Zhao et al., 2014). The algorithm contains two nested loops. In the *outer loop*, we separate all coordinates into two sets: active set and inactive set. Such a partition is based on some heuristic greedy scheme, such as gradient thresholding (also called strong rule, Tibshirani et al. (2012)). Then within each iteration of the middle loop, the *inner loop* only updates coordinates in the active set in a cyclic manner until convergence, where the coordinates in the inactive set remain to be zero. After the inner loop converges, we update the active set based on a greedy selection rule that further decreases the objective value, and repeat the inner loop. Such a procedure continues until the active set no longer changes in the outer loop. We provide the algorithm description as follows and refer Zhao et al. (2014) for further details of active set based coordinate minimization. We use $(p)$ to index the $p$-th iteration of the outer loop, and $(p,l)$ to index the $l$-th iteration of the inner loop at the $p$-th iteration of the outer loop.

**Inner Loop**. The active set $\mathcal{A}$ and inactive set $\mathcal{A}_\perp$ are respectively set as

$$\mathcal{A} \leftarrow \{j \mid \theta_j \neq 0\} = \{j_1, j_2, \ldots, j_s\} \ \text{and} \ \mathcal{A}_\perp \leftarrow \{j \mid j \notin \mathcal{A}\},$$

where $j_1 < j_2 < \ldots < j_s$. A coordinate-wise minimization of (A.2) is performed throughout the inner loop. Specifically, given $\theta^{(p,l)}$ at the $l$-th iteration of the inner loop, we solve (A.2) by only considering the $j$-th coordinate in the active set and fix the rest coordinates in a cyclic manner for all $j = j_1, j_2, \ldots, j_s$, i.e.,

$$\widehat{\Delta\theta_j} = \underset{\Delta\theta_j}{\operatorname{argmin}} \ \frac{1}{2n} \sum_{i=1}^{n} w_i (z_i - \sum_{k \in \mathcal{A}, k \neq j} x_{ik}^\top \Delta\theta_k - x_{ij}^\top \Delta\theta_j)^2 + |\lambda_j(\theta_j + \Delta\theta_j)|. \tag{A.3}$$

Then we update $\theta_j^{(p,l+1)} = \theta_j^{(p,l)} + \widehat{\Delta\theta_j}$. Solving (A.3) has a simple closed form solution by soft thresholding, i.e.,

$$\widehat{\Delta\theta_j} \leftarrow \frac{S(\frac{1}{n}\sum_{i=1}^{n} w_i \delta_{ij}, \lambda_j)}{\frac{1}{n}\sum_{i=1}^{n} w_i x_{ij}^2},$$

where $\delta_{ij} = z_i - \sum_{k \in \mathcal{A}, k \neq j} x_{ik} \Delta\theta_k$ and $S(a,b) = \operatorname{sign}(a)\max\{|a| - b, 0\}$ for real values $a$ and $b$. Moreover, the residual $\delta_{ij}$ can be updated efficiently. Specifically, after the update of $\widehat{\Delta\theta_j}$ for the $j$-th coordinate, then for the next non-zero coordinate, e.g., $j' \in \mathcal{A}$, we update the residual as

$$\delta_{ij'} = \delta_{ij} - x_{ij}^\top \widehat{\Delta\theta_j} + x_{ij'} \Delta\theta_{j'}.$$

This reduces the computational cost of updating each coordinate from $\mathcal{O}(s)$ to $\mathcal{O}(1)$, only with an increase of the memory cost $\mathcal{O}(s)$ for maintaining the previous updates of $\Delta\theta_j$.

Given a convergence parameter $a \in (0,1)$, we terminate the inner loop when

$$\|\theta^{(p,l+1)} - \theta^{(p,l)}\|_2 \leq a\lambda.$$

**Outer Loop.** At the beginning of the outer loop, we initialize the active set $\mathcal{A}^{(0)}$ as follows

$$\mathcal{A}^{(0)} \leftarrow \{j \mid |\nabla_j \mathcal{L}(\theta^{(0)})| \geq (1 - \nu)\lambda\} \cup \{j \mid \theta_j^{(0)} \neq 0\},$$

where $\nabla_j \mathcal{L}(\theta^{(0)})$ is the $j$-th entry of $\nabla \mathcal{L}(\theta^{(0)})$, $\nu \in (0, 0.1)$ is a thresholding parameter, and the inactive set is $\mathcal{A}_\perp^{(0)} = \{j \mid j \notin \mathcal{A}^{(0)}\}$.

Suppose at the $p$-th iteration of the outer loop, the active set is $\mathcal{A}^{(p)}$. We then perform the inner loop introduced above using $\mathcal{A}^{(p)}$ until the convergence of the inner loop and denote $\theta^{(p+1)} = \theta^{(p,l)}$, which is the output of the inner loop. Next, we describe how to update the active set $\mathcal{A}^{(p)}$ using the following greedy selection rule.

- We first shrink the active set as follows. The active coordinate minimization (inner loop) may yield zero solutions on $\mathcal{A}^{(p)}$. We eliminate the zero coordinates of $\theta^{(p+1)}$ from $\mathcal{A}^{(p)}$, and update the intermediate active set and inactive set respectively as

$$\mathcal{A}^{(p+\frac{1}{2})} \leftarrow \{j \in \mathcal{A}^{(p)} \mid \theta_j^{(p+1)} \neq 0\} \text{ and } \mathcal{A}_\perp^{(p+\frac{1}{2})} \leftarrow \{j \mid j \notin \mathcal{A}^{(p+\frac{1}{2})}\}.$$

- We then expand the active set as follows. Denote

$$j^{(p)} = \underset{j \in \mathcal{A}_\perp^{(p+\frac{1}{2})}}{\operatorname{argmax}} |\nabla_j \mathcal{L}(\theta^{(p+1)})|.$$

The outer loop is terminated if

$$|\nabla_{j^{(p)}}\mathcal{L}(\theta^{(p+1)})| \leq (1+\delta)\lambda,$$

where $\delta \ll 1$ is a real positive convergence parameter, e.g., $\delta = 10^{-5}$. Otherwise, we update the sets as

$$\mathcal{A}^{(p+1)} \leftarrow \mathcal{A}^{(p+\frac{1}{2})} \cup \{j^{(p)}\} \text{ and } \mathcal{A}_\perp^{(p+1)} \leftarrow \mathcal{A}_\perp^{(p+\frac{1}{2})} \backslash \{j^{(p)}\},$$

# B  Proof of Intermediate Results for Theorem 3.5

For notational convenience, we denote

$$\mathcal{R}_\lambda^{\ell_1}(\theta) = \|\lambda \odot \theta\|_1.$$

We also introduce an important notion as follows, which is closely related with the SE properties.

**Definition B.1.** *We denote the local $\ell_1$ cone as*

$$\mathcal{C}(s,\vartheta,R) = \left\{ v, \theta : \mathcal{S} \subseteq \mathcal{M}, |\mathcal{M}| \leq s, \|v_{\mathcal{M}_\perp}\|_1 \leq \vartheta \|v_\mathcal{M}\|_1, \|\theta - \theta^*\|_2 \leq R \right\}.$$

*Then we define the largest and smallest **localized restricted eigenvalues** (LRE) as*

$$\psi_{s,\vartheta,R}^+ = \sup_{u,\theta} \left\{ \frac{v^\top \nabla^2 \mathcal{L}(\theta) v}{v^\top v} : (v,\theta) \in \mathcal{C}(s,\vartheta,R) \right\},$$

$$\psi_{s,\vartheta,R}^- = \inf_{u,\theta} \left\{ \frac{v^\top \nabla^2 \mathcal{L}(\theta) v}{v^\top v} : (v,\theta) \in \mathcal{C}(s,\vartheta,R) \right\}.$$

The following proposition demonstrates the relationships between SE and LRE. The proof can be found in Bühlmann and Van De Geer (2011), thus is omitted here.

**Proposition B.2.** *Given any $\theta, \theta' \in \mathcal{C}(s,\vartheta,R) \cap \mathcal{B}(\theta^*,R)$, we have*

$$c_1 \psi_{s,\vartheta,R}^- \leq \rho_s^- \leq c_2 \psi_{s,\vartheta,R}^-, \text{ and } c_3 \psi_{s,\vartheta,R}^+ \leq \rho_s^+ \leq c_4 \psi_{s,\vartheta,R}^+.$$

*where $c_1$, $c_2$, $c_3$, and $c_4$ are constants.*

## B.1  Proof of Lemma 6.4

We first demonstrate the sparsity of the update. For notational convenience, we omit the stage index $\{K\}$. Since $\theta^{(t+1)}$ is the minimizer to the proximal Newton problem, we have

$$\nabla^2 \mathcal{L}(\theta^{(t)})(\theta^{(t+1)} - \theta^{(t)}) + \nabla \mathcal{L}(\theta^{(t)}) + \lambda \odot \xi^{(t+1)} = 0,$$

where $\xi^{(t+1)} \in \partial \|\theta^{(t+1)}\|_1$.

It follows from Fan et al. (2015) that if Assumption 3 holds, then we have $\min_{j \in S'_\perp}\{\lambda_j\} \geq \lambda_{\text{tgt}}/2$ for some set $S' \supset S$ with $|S'| \leq 2s^*$. Then the analysis of sparsity of can be performed through $\lambda_{\text{tgt}}$ directly.

We then consider the following decomposition

$$\nabla^2 \mathcal{L}(\theta^{(t)})(\theta^{(t+1)} - \theta^{(t)}) + \nabla \mathcal{L}(\theta^{(t)})$$
$$= \underbrace{\nabla^2 \mathcal{L}(\theta^{(t)})(\theta^{(t+1)} - \theta^*)}_{V_1} + \underbrace{\nabla^2 \mathcal{L}(\theta^{(t)})(\theta^* - \theta^{(t)})}_{V_2} + \underbrace{\nabla \mathcal{L}(\theta^{(t)}) - \nabla \mathcal{L}(\theta^*)}_{V_3} + \underbrace{\nabla \mathcal{L}(\theta^*)}_{V_4}.$$

Consider the following sets:

$$\mathcal{A}_i = \left\{ j \in \mathcal{S}'_\perp \; : \; |(V_i)_j| \geq \lambda_{\text{tgt}}/4 \right\}, \text{ for all } i \in \{1, 2, 3, 4\}.$$

**Set $\mathcal{A}_2$.** We have $\mathcal{A}_2 = \left\{ j \in \mathcal{S}'_\perp : |(\nabla^2 \mathcal{L}(\theta^{(t)})(\theta^* - \theta^{(t)}))_j| \geq \lambda_{\text{tgt}}/4 \right\}$. Consider a subset $\mathcal{S}' \subset \mathcal{A}_2$ with $|\mathcal{S}'| = s' \leq \widetilde{s}$. Suppose we choose a vector $v \in \mathbb{R}^d$ such that $\|v\|_\infty = 1$ and $\|v\|_0 = s'$ with $s' \lambda_{\text{tgt}}/4 \leq v^\top \nabla^2 \mathcal{L}(\theta^{(t)})(\theta^* - \theta^{(t)})$. Then we have

$$s'\lambda_{\text{tgt}}/4 \leq v^\top \nabla^2 \mathcal{L}(\theta^{(t)})(\theta^* - \theta^{(t)}) \leq \|v(\nabla^2 \mathcal{L}(\theta^{(t)}))^{\frac{1}{2}}\|_2 \|(\nabla^2 \mathcal{L}(\theta^{(t)}))^{\frac{1}{2}}(\theta^* - \theta^{(t)})\|_2$$
$$\overset{(i)}{\leq} \sqrt{\rho^+_{s^*+2\widetilde{s}} \rho^+_{s'}} \|v\|_2 \|\theta^* - \theta^{(t)}\|_2 \overset{(ii)}{\leq} \sqrt{s'\rho^+_{s^*+2\widetilde{s}} \rho^+_{s'}} \|\theta^* - \theta^{(t)}\|_2$$
$$\overset{(iii)}{\leq} \frac{C'\sqrt{s'\rho^+_{s^*+2\widetilde{s}} \rho^+_{s'}} \lambda_{\text{tgt}} \sqrt{s^*}}{\rho^-_{s^*+2\widetilde{s}}}, \tag{B.1}$$

where $(i)$ is from the SE properties, $(ii)$ is from the definition of $v$, and $(iii)$ is from $\|\theta^{(t)} - \theta^*\|_2 \leq C'\lambda_{\text{tgt}}\sqrt{s^*}/\rho^-_{s^*+2\widetilde{s}}$. Then (B.1) implies

$$s' \leq \frac{C_2 \rho^+_{s^*+2\widetilde{s}} \rho^+_{s'} s^*}{(\rho^-_{s^*+2\widetilde{s}})^2} \leq C_2 \kappa^2_{s^*+2\widetilde{s}} s^*, \tag{B.2}$$

where the last inequality is from the fact that $s' = |\mathcal{S}'|$ achieves the maximum possible value such that $s' \leq \widetilde{s}$ for any subset $\mathcal{S}'$ of $\mathcal{A}_2$. (B.2) implies that $s' < \widetilde{s}$, so wo must have $\mathcal{S}' = \mathcal{A}_2$ to attain the maximum. Then we have

$$|\mathcal{A}_2| = s' \leq C_2 \kappa^2_{s^*+2\widetilde{s}} s^*.$$

**Set $\mathcal{A}_3$.** We have $\mathcal{A}_3 = \left\{ j \in \mathcal{S}'_\perp : \left|(\nabla \mathcal{L}(\theta^{(t)}) - \nabla \mathcal{L}(\theta^*))_i\right| \geq \lambda_{\text{tgt}}/4 \right\}$. Suppose we choose a vector $v \in \mathbb{R}^d$ such that $\|v\|_\infty = 1$, $\|v\|_0 = |\mathcal{A}_3|$ and

$$v^\top \left( \nabla \mathcal{L}(\theta^{(t)}) - \nabla \mathcal{L}(\theta^*) \right) = \sum_{i \in \mathcal{A}_3} v_i \left( \nabla \mathcal{L}(\theta^{(t)}) - \nabla \mathcal{L}(\theta^*) \right)_i = \sum_{i \in \mathcal{A}_3} \left| \left( \nabla \mathcal{L}(\theta^{(t)}) - \nabla \mathcal{L}(\theta^*) \right)_i \right| \geq \lambda_{\text{tgt}} |\mathcal{A}_3|/4. \tag{B.3}$$

Then we have

$$v^\top \left( \nabla \mathcal{L}(\theta^{(t)}) - \nabla \mathcal{L}(\theta^*) \right) \leq \|v\|_2 \|\nabla \mathcal{L}(\theta^{(t)}) - \nabla \mathcal{L}(\theta^*)\|_2 \overset{(i)}{\leq} \sqrt{|\mathcal{A}_3|} \cdot \|\nabla \mathcal{L}(\theta^{(t)}) - \nabla \mathcal{L}(\theta^*)\|_2$$
$$\overset{(ii)}{\leq} \rho^+_{s^*+2\widetilde{s}} \sqrt{|\mathcal{A}_3|} \cdot \|\theta^{(t)} - \theta^*\|_2, \tag{B.4}$$

where $(i)$ is from the definition of $v$, and $(ii)$ is from the mean value theorem and analogous argument for $\mathcal{A}_2$.

Combining (B.3) and (B.4), we have

$$\lambda_{\text{tgt}}|\mathcal{A}_3| \leq 4\rho^+_{s^*+2\widetilde{s}}\sqrt{|\mathcal{A}_3|} \cdot \|\theta - \theta^*\|_2 \overset{(i)}{\leq} 8\lambda_{\text{tgt}}\kappa_{s^*+2\widetilde{s}}\sqrt{3s^*|\mathcal{A}_3|}$$

where $(i)$ is from $\|\theta^{(t)} - \theta^*\|_2 \leq C'\lambda_{\text{tgt}}\sqrt{s^*}/\rho^-_{s^*+2\widetilde{s}}$ and definition of $\kappa_{s^*+2\widetilde{s}} = \rho^+_{s^*+2\widetilde{s}}/\rho^-_{s^*+2\widetilde{s}}$. This implies

$$|\mathcal{A}_3| \leq C_3 \kappa^2_{s^*+2\widetilde{s}} s^*.$$

**Set $\mathcal{A}_4$.** By Assumption 3 and $\lambda_{\text{tgt}} \geq 4\|\nabla \mathcal{L}(\theta^*)\|_\infty$, we have

$$0 \leq |V_4| \leq \sum_{i \in \mathcal{S}^*_\perp} \frac{4}{\lambda_{\text{tgt}}} |(\nabla \mathcal{L}(\theta^*))_i| \cdot \mathbb{1}(|(\nabla \mathcal{L}(\theta^*))_i| > \lambda_{\text{tgt}}/(4)) = \sum_{i \in \mathcal{S}^*_\perp} \frac{4}{\lambda_{\text{tgt}}} |(\nabla \mathcal{L}(\theta^*))_i| \cdot 0 = 0, \quad (B.5)$$

**Set $A_1$.** From Lemma E.1, we have $\mathcal{F}_\lambda(\theta^{(t+1)}) \leq \mathcal{F}_\lambda(\theta^*) + \frac{\lambda_{\text{tgt}}}{4}\|\theta^{(t+1)} - \theta^*\|_1$. This implies

$$\begin{aligned}
\mathcal{L}(\theta^{(t+1)}) - \mathcal{L}(\theta^*) &\leq \lambda_{\text{tgt}}(\|\theta^*\|_1 - \|\theta^{(t+1)}\|_1) + \frac{\lambda_{\text{tgt}}}{4}\|\theta^{(t+1)} - \theta^*\|_1 \\
&= \lambda_{\text{tgt}}(\|\theta^*_{\mathcal{S}'}\|_1 - \|\theta^{(t+1)}_{\mathcal{S}'}\|_1 - \|\theta^{(t+1)}_{\mathcal{S}'_\perp}\|_1) + \frac{\lambda_{\text{tgt}}}{4}\|\theta^{(t+1)} - \theta^*\|_1 \\
&\leq \frac{5\lambda_{\text{tgt}}}{4}\|\theta^{(t+1)}_{\mathcal{S}'} - \theta^*_{\mathcal{S}'}\|_1 - \frac{3\lambda_{\text{tgt}}}{4}\|\theta^{(t+1)}_{\mathcal{S}'_\perp} - \theta^*_{\mathcal{S}'_\perp}\|_1.
\end{aligned} \quad (B.6)$$

where the equality holds since $\theta^*_{\mathcal{S}'_\perp} = 0$. On the other hand, we have

$$\begin{aligned}
\mathcal{L}(\theta^{(t+1)}) - \mathcal{L}(\theta^*) &\overset{(i)}{\geq} \nabla \mathcal{L}(\theta^*)(\theta^{(t+1)} - \theta^*) \geq -\|cL(\theta^*)\|_\infty \|\theta^{(t+1)} - \theta^*\|_1 \\
&\overset{(ii)}{\geq} -\frac{\lambda_{\text{tgt}}}{4}\|\theta^{(t+1)} - \theta^*\|_1 = -\frac{\lambda_{\text{tgt}}}{4}\|\theta^{(t+1)}_{\mathcal{S}'} - \theta^*_{\mathcal{S}'}\|_1 - \frac{\lambda_{\text{tgt}}}{4}\|\theta^{(t+1)}_{\mathcal{S}'_\perp} - \theta^*_{\mathcal{S}'_\perp}\|_1,
\end{aligned} \quad (B.7)$$

where $(i)$ is from the convexity of $\mathcal{L}$ and $(ii)$ is from Assumption 3. Combining (B.6) and (B.7), we have

$$\|\theta^{(t+1)}_{\mathcal{S}'_\perp} - \theta^*_{\mathcal{S}'_\perp}\|_1 \leq 3\|\theta^{(t+1)}_{\mathcal{S}'} - \theta^*_{\mathcal{S}'}\|_1,$$

which implies that $\theta^{(t+1)} - \theta^* \in \mathcal{C}(s^*, 3, R)$ with respect to the set $\mathcal{S}'$.

We have $\mathcal{A}_4 = \{j \in \mathcal{S}'_\perp : |(\nabla^2 \mathcal{L}(\theta^{(t)})(\theta^* - \theta^{(t+1)}))_j| \geq \lambda_{\text{tgt}}/4\}$. Consider a subset $\mathcal{S}' \subset \mathcal{A}_2$ with $|\mathcal{S}'| = s' \leq \widetilde{s}$ and a vector $v \in \mathbb{R}^d$ similar to that in $\mathcal{A}_2$. Then we have

$$\begin{aligned}
s'\lambda_{\text{tgt}}/4 &\leq v^\top \nabla^2 \mathcal{L}(\theta^{(t)})(\theta^{(t+1)} - \theta^*) \leq \|v(\nabla^2 \mathcal{L}(\theta^{(t)}))^{\frac{1}{2}}\|_2 \|(\nabla^2 \mathcal{L}(\theta^{(t)}))^{\frac{1}{2}}(\theta^{(t+1)} - \theta^*)\|_2 \\
&\overset{(i)}{\leq} c_1\sqrt{\rho^+_{s^*+2\widetilde{s}}\rho^+_{s'}} \|v\|_2 \|\theta^* - \theta^{(t+1)}\|_2 \overset{(ii)}{\leq} c_1\sqrt{s'\rho^+_{s^*+2\widetilde{s}}\rho^+_{s'}} \|\theta^* - \theta^{(t+1)}\|_2 \\
&\overset{(iii)}{\leq} \frac{c_2\sqrt{s'\rho^+_{s^*+2\widetilde{s}}\rho^+_{s'}}\lambda_{\text{tgt}}\sqrt{s^*}}{\rho^-_{s^*+2\widetilde{s}}},
\end{aligned} \quad (B.8)$$

where $(i)$ is from SE condition and Proposition B.2, $(ii)$ is from the definition of $v$, and $(iii)$ is from $\|\theta^{(t+1)} - \theta^*\|_2 \leq C'\lambda_{\text{tgt}}\sqrt{s^*}/\rho^-_{s^*+2\widetilde{s}}$. Following analogous argument in for $A_2$, we have

$$|A_1| \leq C_1 \kappa^2_{s^*+2\widetilde{s}} s^*.$$

Combining the results for Set $A_1 \sim A_4$, we have that there exists some constant $C_0$ such that

$$\|\theta^{(t+\frac{1}{2})}_{\mathcal{S}_\perp}\|_0 \leq C_0 \kappa^2_{s^*+2\widetilde{s}} s^* \leq \widetilde{s}.$$

From Lemma 6.6, we further have that the step size satisfies $\eta_t = 1$, then we have $\theta^{(t+1)} = \theta^{(t+\frac{1}{2})}$. The estimation error follows directly from Lemma E.2.

## B.2 Proof of Lemma 6.5

For notational simplicity, we introduce the following proximal operator,

$$\text{prox}^{H,g}_r(\theta) = \text{argmin}_{\theta'} r(\theta') + g^\top(\theta' - \theta) + \frac{1}{2}\|\theta' - \theta\|^2_H.$$

Then we have

$$\theta^{(t+1)} = \text{prox}^{\nabla^2 \mathcal{L}(\theta^{(t)}), \nabla \mathcal{L}(\theta^{(t)})}_{\mathcal{R}^{\ell_1}_{\lambda^{\{K\}}}(\theta^{(t)})}\left(\theta^{(t)}\right).$$

By Lemma 6.4, we have

$$\|\theta^{(t+1)}_{\mathcal{S}_\perp}\|_0 \leq \widetilde{s}.$$

By the KKT condition of function $\min \mathcal{F}_{\lambda^{\{K\}}}$, i.e., $-\nabla \mathcal{L}(\overline{\theta}^{\{K\}}) \in \partial \mathcal{R}^{\ell_1}_{\lambda^{\{K\}}}(\overline{\theta}^{\{K\}})$, we also have

$$\overline{\theta}^{\{K\}} = \text{prox}^{\nabla^2 \mathcal{L}(\theta^{(t)}), \nabla \mathcal{L}(\overline{\theta}^{\{K\}})}_{\mathcal{R}^{\ell_1}_{\lambda^{\{K\}}}(\overline{\theta}^{\{K\}})}\left(\overline{\theta}^{\{K\}}\right).$$

By monotonicity of sub-gradient of a convex function, we have the *strictly non-expansive* property: for any $\theta, \theta' \in \mathbb{R}$, let $u = \text{prox}^{H,g}_r(\theta)$ and $v = \text{prox}^{H,g'}_r(\theta')$, then

$$(u-v)^\top H(\theta - \theta') - (u-v)^\top(g - g') \geq \|u-v\|^2_H.$$

Thus by the strictly non-expansive property of the proximal operator, we obtain

$$\|\theta^{(t+1)} - \overline{\theta}^{\{K\}}\|^2_{\nabla^2 \mathcal{L}(\overline{\theta}^{\{K\}})} \leq \left(\theta^{(t+1)} - \overline{\theta}^{\{K\}}\right)^\top \left[\nabla^2 \mathcal{L}(\theta^{(t)})\left(\theta^{(t)} - \overline{\theta}^{\{K\}}\right) + \left(\nabla \mathcal{L}(\overline{\theta}^{\{K\}}) - \nabla \mathcal{L}(\theta^{(t)})\right)\right]$$

$$\leq \|\theta^{(t+1)} - \overline{\theta}^{\{K\}}\|_2 \left\|\nabla^2 \mathcal{L}(\theta^{(t)})\left(\theta^{(t)} - \overline{\theta}^{\{K\}}\right) + \left(\nabla \mathcal{L}(\overline{\theta}^{\{K\}}) - \nabla \mathcal{L}(\theta^{(t)})\right)\right\|_2. \quad \text{(B.9)}$$

Note that both $\|\theta^{(t+1)}\|_0 \leq \widetilde{s}$ and $\|\overline{\theta}^{\{K\}}\|_0 \leq \widetilde{s}$. On the other hand, from the SE properties, we have

$$\|\theta^{(t+1)} - \overline{\theta}^{\{K\}}\|^2_{\nabla^2 \mathcal{L}(\overline{\theta}^{\{K\}})} = (\theta^{(t+1)} - \overline{\theta}^{\{K\}})^\top \nabla^2 \mathcal{L}(\overline{\theta}^{\{K\}})(\theta^{(t+1)} - \overline{\theta}^{\{K\}}) \geq \rho^-_{s^*+2\widetilde{s}}\|\theta^{(t+1)} - \overline{\theta}^{\{K\}}\|^2_2. \quad \text{(B.10)}$$

Combining (B.9) and (B.10), we have

$$\left\|\theta^{(t+1)} - \overline{\theta}^{\{K\}}\right\|_2 \leq \frac{1}{\rho^-_{s^*+2\widetilde{s}}} \left\|\nabla^2\mathcal{L}(\theta^{(t)})\left(\theta^{(t)} - \overline{\theta}^{\{K\}}\right) + \left(\nabla\mathcal{L}(\overline{\theta}^{\{K\}}) - \nabla\mathcal{L}(\theta^{(t)})\right)\right\|_2$$

$$= \frac{1}{\rho^-_{s^*+2\widetilde{s}}} \left\|\int_0^1 \left[\nabla^2\mathcal{L}\left(\theta^{(t)} + \tau\left(\overline{\theta}^{\{K\}} - \theta^{(t)}\right)\right) - \nabla^2\mathcal{L}\left(\theta^{(t)}\right)\right] \cdot \left(\overline{\theta}^{\{K\}} - \theta^{(t)}\right) d\tau\right\|_2$$

$$\leq \frac{1}{\rho^-_{s^*+2\widetilde{s}}} \int_0^1 \left\|\left[\nabla^2\mathcal{L}\left(\theta^{(t)} + \tau\left(\overline{\theta}^{\{K\}} - \theta^{(t)}\right)\right) - \nabla^2\mathcal{L}\left(\theta^{(t)}\right)\right] \cdot \left(\overline{\theta}^{\{K\}} - \theta^{(t)}\right)\right\|_2 d\tau$$

$$\leq \frac{L_{s^*+2\widetilde{s}}}{2\rho^-_{s^*+2\widetilde{s}}} \left\|\theta^{(t)} - \overline{\theta}^{\{K\}}\right\|_2^2,$$

where the last inequality is from the local restricted Hessian smoothness of $\mathcal{L}$. Then we finish the proof by the definition of $R$.

## B.3 Proof of Lemma 6.6

Suppose the step size $\eta_t < 1$. Note that we do not need the step size to be $\eta_t = 1$ in Lemma 6.4 and Lemma 6.5. We denote $\Delta\theta^{(t)} = \theta^{(t+\frac{1}{2})} - \theta^{(t)}$. Then we have

$$\left\|\Delta\theta^{(t)}\right\|_2 \overset{(i)}{\leq} \left\|\theta^{(t)} - \overline{\theta}^{\{K\}}\right\|_2 + \left\|\theta^{(t+\frac{1}{2})} - \overline{\theta}^{\{K\}}\right\|_2 \overset{(ii)}{\leq} \left\|\theta^{(t)} - \overline{\theta}^{\{K\}}\right\|_2 + \frac{L_{s^*+2\widetilde{s}}}{2\rho^-_{s^*+2\widetilde{s}}} \left\|\theta^{(t)} - \overline{\theta}^{\{K\}}\right\|_2^2$$

$$\overset{(iii)}{\leq} \frac{3}{2}\left\|\theta^{(t)} - \overline{\theta}^{\{K\}}\right\|_2, \tag{B.11}$$

where $(i)$ is from triangle inequality, $(ii)$ is from Lemma 6.5, and $(iii)$ is from $\left\|\theta^{(t)} - \overline{\theta}^{\{K\}}\right\|_2 \leq R \leq \frac{\rho^-_{s^*+2\widetilde{s}}}{L_{s^*+2\widetilde{s}}}$.

By Lemma 6.4, we have

$$\left\|\Delta\theta^{(t)}_{\mathcal{S}_\perp}\right\|_0 \leq 2\widetilde{s}.$$

To show $\eta_t = 1$, it is now suffice to demonstrate that

$$\mathcal{F}_{\lambda^{\{K\}}}(\theta^{(t+\frac{1}{2})}) - \mathcal{F}_{\lambda^{\{K\}}}(\theta^{(t)}) \leq \frac{1}{4}\gamma_t.$$

By expanding $\mathcal{F}_{\lambda^{\{K\}}}$, we have

$$\mathcal{F}_{\lambda^{\{K\}}}(\theta^{(t)} + \Delta\theta^{(t)}) - \mathcal{F}_{\lambda^{\{K\}}}(\theta^{(t)}) = \mathcal{L}(\theta^{(t)} + \Delta\theta^{(t)}) - \mathcal{L}(\theta^{(t)}) + \mathcal{R}^{\ell_1}_{\lambda^{\{K\}}}(\theta^{(t)} + \Delta\theta^{(t)}) - \mathcal{R}^{\ell_1}_{\lambda^{\{K\}}}(\theta^{(t)})$$

$$\overset{(i)}{\leq} \nabla\mathcal{L}(\theta^{(t)})^\top \Delta\theta^{(t)} + \frac{1}{2}\Delta(\theta^{(t)})^\top \nabla^2\mathcal{L}(\theta)\Delta\theta^{(t)} + \frac{L_{s^*+2\widetilde{s}}}{6}\left\|\Delta\theta^{(t)}\right\|_2^3 + \mathcal{R}^{\ell_1}_{\lambda^{\{K\}}}(\theta^{(t)} + \Delta\theta^{(t)}) - \mathcal{R}^{\ell_1}_{\lambda^{\{K\}}}(\theta^{(t)})$$

$$\overset{(ii)}{\leq} \gamma_t - \frac{1}{2}\gamma_t + \frac{L_{s^*+2\widetilde{s}}}{6}\left\|\Delta\theta^{(t)}\right\|_2^3 \overset{(iii)}{\leq} \frac{1}{2}\gamma_t + \frac{L_{s^*+2\widetilde{s}}}{6\rho^-_{s^*+2\widetilde{s}}}\left\|\Delta\theta^{(t)}\right\|_{\nabla^2\mathcal{L}(\theta)}\left\|\Delta\theta^{(t)}\right\|_2$$

$$\overset{(iv)}{\leq} \left(\frac{1}{2} - \frac{L_{s^*+2\widetilde{s}}}{6\rho^-_{s^*+2\widetilde{s}}}\left\|\Delta\theta^{(t)}\right\|_2\right)\gamma_t \overset{(v)}{\leq} \frac{1}{4}\gamma_t,$$

where $(i)$ is from the restricted Hessian smooth condition, $(ii)$ and $(iv)$ are from Lemma 6.7, $(iii)$ is from the same argument of (B.10), and $(v)$ is from (B.11), $\gamma_t < 0$, and $\left\|\theta^{(t)} - \overline{\theta}^{\{K\}}\right\|_2 \leq R \leq \frac{\rho^-_{s^*+2\widetilde{s}}}{L_{s^*+2\widetilde{s}}}$. This implies $\theta^{(t+1)} = \theta^{(t+\frac{1}{2})}$.

## B.4 Proof of Lemma 6.7

We denote $H = \nabla^2 \mathcal{L}(\theta^{(t)})$. Since $\Delta\theta^{(t)}$ is the solution for

$$\min_{\Delta\theta^{(t)}} \nabla\mathcal{L}\left(\theta^{(t)}\right)^\top \cdot \Delta\theta^{(t)} + \frac{1}{2}\left\|\Delta\theta^{(t)}\right\|_H^2 + \mathcal{R}_{\lambda^{\{K\}}}^{\ell_1}\left(\theta^{(t)} + \Delta\theta^{(t)}\right)$$

then for any $\eta_t \in (0, 1]$, we have

$$\eta_t \nabla\mathcal{L}\left(\theta^{(t)}\right)^\top \cdot \Delta\theta^{(t)} + \frac{\eta_t^2}{2}\left\|\Delta\theta^{(t)}\right\|_H^2 + \mathcal{R}_{\lambda^{\{K\}}}^{\ell_1}\left(\theta^{(t)} + \eta_t\Delta\theta^{(t)}\right)$$
$$\geq \nabla\mathcal{L}\left(\theta^{(t)}\right)^\top \cdot \Delta\theta^{(t)} + \frac{1}{2}\left\|\Delta\theta^{(t)}\right\|_H^2 + \mathcal{R}_{\lambda^{\{K\}}}^{\ell_1}\left(\theta^{(t)} + \Delta\theta^{(t)}\right)$$

By the convexity of $\mathcal{R}_{\lambda^{\{K\}}}^{\ell_1}$, we have

$$\eta_t \nabla\mathcal{L}\left(\theta^{(t)}\right)^\top \cdot \Delta\theta^{(t)} + \frac{\eta_t^2}{2}\left\|\Delta\theta^{(t)}\right\|_H^2 + \eta_t\mathcal{R}_{\lambda^{\{K\}}}^{\ell_1}\left(\theta^{(t)} + \Delta\theta^{(t)}\right) + (1-\eta_t)\mathcal{R}_{\lambda^{\{K\}}}^{\ell_1}(\theta^{(t)})$$
$$\geq \nabla\mathcal{L}\left(\theta^{(t)}\right)^\top \cdot \Delta\theta^{(t)} + \frac{1}{2}\left\|\Delta\theta^{(t)}\right\|_H^2 + \mathcal{R}_{\lambda^{\{K\}}}^{\ell_1}\left(\theta^{(t)} + \Delta\theta^{(t)}\right).$$

Rearranging the terms, we obtain

$$(1-\eta_t)\left(\nabla\mathcal{L}\left(\theta^{(t)}\right)^\top \cdot \Delta\theta^{(t)} + \mathcal{R}_{\lambda^{\{K\}}}^{\ell_1}\left(\theta^{(t)} - \Delta\theta^{(t)}\right) - \mathcal{R}_{\lambda^{\{K\}}}^{\ell_1}(\theta^{(t)})\right) + \frac{1-\eta_t^2}{2}\left\|\Delta\theta^{(t)}\right\|_H^2 \leq 0$$

Canceling the $(1-\eta_t)$ factor from both sides and let $\eta_t \to 1$, we obtain the desired inequality,

$$\gamma_t \leq -\left\|\Delta\theta^{(t)}\right\|_H^2.$$

## B.5 Proof of Lemma 6.8

We first demonstrate an upper bound of the approximate KKT parameter $\omega_{\lambda^{\{K\}}}$. Given the solution $\theta^{(t-1)}$ from the $(t-1)$-th iteration, the optimal solution at $t$-th iteration satisfies the KKT condition:

$$\nabla^2\mathcal{L}(\theta^{(t-1)})(\theta^{(t)} - \theta^{(t-1)}) + \nabla\mathcal{L}(\theta^{(t-1)}) + \lambda^{\{K\}} \odot \xi^{(t)} = 0,$$

where $\xi^{(t)} \in \partial\|\theta^{(t)}\|_1$. Then for any vector $v$ with $\|v\|_2 \leq \|v\|_1 = 1$ and $\|v\|_0 \leq s^* + 2\widetilde{s}$, we have

$$(\nabla\mathcal{L}(\theta^{(t)}) + \lambda^{\{K\}} \odot \xi^{(t)})^\top v = (\nabla\mathcal{L}(\theta^{(t)}))^\top v - (\nabla^2\mathcal{L}(\theta^{(t-1)})(\theta^{(t)} - \theta^{(t-1)}) + \nabla\mathcal{L}(\theta^{(t-1)}))^\top v$$
$$= (\nabla\mathcal{L}(\theta^{(t)}) - \nabla\mathcal{L}(\theta^{(t-1)}))^\top v - (\nabla^2\mathcal{L}(\theta^{(t-1)})(\theta^{(t)} - \theta^{(t-1)}))^\top v$$
$$\overset{(i)}{\leq} \left\|(\nabla^2\mathcal{L}(\widetilde{\theta}))^{\frac{1}{2}}(\theta^{(t)} - \theta^{(t-1)})\right\|_2 \cdot \left\|v^\top(\nabla^2\mathcal{L}(\widetilde{\theta}))^{\frac{1}{2}}\right\|_2$$
$$+ \left\|(\nabla^2\mathcal{L}(\theta^{(t-1)}))^{\frac{1}{2}}(\theta^{(t)} - \theta^{(t-1)})\right\|_2 \cdot \left\|v^\top(\nabla^2\mathcal{L}(\theta^{(t-1)}))^{\frac{1}{2}}\right\|_2$$
$$\overset{(ii)}{\leq} 2\rho_{s^*+2\widetilde{s}}^+\left\|\theta^{(t)} - \theta^{(t-1)}\right\|_2, \qquad \text{(B.12)}$$

where (i) is from mean value theorem with some $\widetilde{\theta} = (1-a)\theta^{(t-1)} + a\theta^{(t)}$ for some $a \in [0,1]$ and Cauchy-Schwarz inequality, and (ii) is from the SE properties. Take the supremum of the L.H.S. of (B.12) with respect to $v$, we have

$$\left\|\nabla \mathcal{L}(\theta^{(t)}) + \lambda^{\{K\}} \odot \xi^{(t)}\right\|_\infty \leq 2\rho^+_{s^*+2\widetilde{s}} \left\|\theta^{(t)} - \theta^{(t-1)}\right\|_2. \tag{B.13}$$

Then from Lemma 6.5, we have

$$\left\|\theta^{(t+1)} - \overline{\theta}^{\{K\}}\right\|_2 \leq \left(\frac{L_{s^*+2\widetilde{s}}}{2\rho^-_{s^*+2\widetilde{s}}}\right)^{1+2+4+\ldots+2^{t-1}} \left\|\theta^{(0)} - \overline{\theta}^{\{K\}}\right\|_2^{2^\top} \leq \left(\frac{L_{s^*+2\widetilde{s}}}{2\rho^-_{s^*+2\widetilde{s}}} \left\|\theta^{(0)} - \overline{\theta}^{\{K\}}\right\|_2\right)^{2^t}.$$

By (B.13) and (B.11) by taking $\Delta\theta^{(t-1)} = \theta^{(t)} - \theta^{(t-1)}$, we obtain

$$\omega_{\lambda^{\{K\}}}\left(\theta^{(t)}\right) \leq 2\rho^+_{s^*+2\widetilde{s}} \left\|\theta^{(t)} - \theta^{(t-1)}\right\|_2 \leq 3\rho^+_{s^*+2\widetilde{s}} \left\|\theta^{(t-1)} - \overline{\theta}^{\{K\}}\right\|_2 \leq 3\rho^+_{s^*+2\widetilde{s}} \left(\frac{L_{s^*+2\widetilde{s}}}{2\rho^-_{s^*+2\widetilde{s}}} \left\|\theta^{(0)} - \overline{\theta}^{\{K\}}\right\|_2\right)^{2^t}.$$

By requiring the R.H.S. equal to $\varepsilon$ we obtain

$$t = \log \frac{\log\left(\frac{3\rho^+_{s^*+2\widetilde{s}}}{\varepsilon}\right)}{\log\left(\frac{2\rho^-_{s^*+2\widetilde{s}}}{L_{s^*+2\widetilde{s}} \left\|\theta^{(0)} - \overline{\theta}^{\{K\}}\right\|_2}\right)} = \log\log\left(\frac{3\rho^+_{s^*+2\widetilde{s}}}{\varepsilon}\right) - \log\log\left(\frac{2\rho^-_{s^*+2\widetilde{s}}}{L_{s^*+2\widetilde{s}} \left\|\theta^{(0)} - \overline{\theta}^{\{K\}}\right\|_2}\right)$$

$$\overset{(i)}{\leq} \log\log\left(\frac{3\rho^+_{s^*+2\widetilde{s}}}{\varepsilon}\right) - \log\log 4 \leq \log\log\left(\frac{3\rho^+_{s^*+2\widetilde{s}}}{\varepsilon}\right),$$

where (i) is from the fact that $\left\|\theta^{(0)} - \overline{\theta}^{\{K\}}\right\|_2 \leq R = \frac{\rho^-_{s^*+2\widetilde{s}}}{2L_{s^*+2\widetilde{s}}}$.

## C Proof of Intermediate Results for Theorem 3.3

### C.1 Proof of Lemma 6.1

Given the assumptions, we will show that for all large enough $t$, we have

$$\|\theta^{(t+1)}_{\mathcal{S}_\perp}\|_0 \leq \widetilde{s}.$$

Following the analysis of Lemma 6.6, Lemma 6.7, and Appendix F, we have that the objective $\mathcal{F}_{\lambda^{\{1\}}}$ has sufficient descendant in each iteration of proximal Newton step, which is also discussed in Yen et al. (2014). Then there exists a constant $T$ such that for all $t \geq T$, we have

$$\mathcal{F}_{\lambda^{\{1\}}}(\theta^{(t)}) \leq \mathcal{F}_{\lambda^{\{1\}}}(\theta^*) + \frac{\lambda_{\text{tgt}}}{4} \|\theta^{(t)} - \theta^*\|_1,$$

where $\|\theta^{(t)} - \theta^*\|_1 \leq c\lambda_{\text{tgt}} \sqrt{s^*}/\rho^-_{s^*+\widetilde{s}}$ from similar analysis in Fan et al. (2015). The rest of the analysis is analogous to that of Lemma 6.4, from which we have $\|\theta^{(t)}_{\mathcal{S}_\perp}\|_0 \leq \widetilde{s}$.

## C.2 Proof of Lemma 6.2

The estimation error is derived analogously from Fan et al. (2015), thus we omit it here. The claim of the quadratic convergence follows directly from Lemma 6.5 given sparse solutions.

## C.3 Proof of Lemma 6.3

The upper bound of the number of iterations for proximal Newton update is obtained by combining Lemma 6.1 and Lemma 6.8. Note that

$$T_1 \leq \log \frac{\log\left(\frac{3\rho^+_{s^*+2\bar{s}}}{\varepsilon}\right)}{\log\left(\frac{2\rho^-_{s^*+2\bar{s}}}{L_{s^*+2\bar{s}}\left\|\theta^{(T+1)}-\overline{\theta}^{\{1\}}\right\|_2}\right)}.$$

Then we obtain the result from $\left\|\theta^{(T+1)}-\overline{\theta}^{\{1\}}\right\|_2 \leq R = \frac{\rho^-_{s^*+2\bar{s}}}{2L_{s^*+2\bar{s}}}$.

# D Proof of Theorem 3.7

It is demonstrated in Ning et al. (2014) that Assumptions 1 ~ 3 hold given the LRE properties defined in Definition B.1. Thus, combining the analyses in Ning et al. (2014) and Proposition B.2, we have that Assumptions 1 ~ 3 hold with high probability. Assumption 4 also holds trivially by choosing $\varepsilon = \frac{c}{\sqrt{n}}$ for some generic constant $c$. The rest of the results follow directly from Theorem 3.5 and the analyses in Zhang (2010b).

# E Further Intermediate Results

**Lemma E.1.** *Given* $\omega_{\lambda^{\{K\}}}(\widehat{\theta}^{\{K\}}) \leq \frac{\lambda_{\text{tgt}}}{8}$, *we have that for all* $t \geq 1$ *at the* $\{K+1\}$*-th stage,*

$$\omega_{\lambda^{\{K+1\}}}(\theta^{(t)}) \leq \frac{\lambda_{\text{tgt}}}{4} \quad \text{and} \quad \mathcal{F}_{\lambda^{\{K+1\}}}(\theta^{(t)}) \leq \mathcal{F}_{\lambda^{\{K+1\}}}(\theta^*) + \frac{\lambda_{\text{tgt}}}{4}\|\theta^{(t)}-\theta^*\|_1.$$

*Proof.* Note that at the $\{K+1\}$-th stage, $\theta^{(0)} = \widehat{\theta}^{\{K\}}$. Then we have

$$\omega_{\lambda^{\{K+1\}}}(\theta^{(0)}) = \min_{\xi \in \partial\|\theta^{(0)}\|_1} \|\nabla\mathcal{L}(\theta^{(0)}) + \lambda^{\{K+1\}} \odot \xi\|_\infty$$

$$\overset{(i)}{\leq} \min_{\xi \in \|\theta^{(0)}\|_1} \|\nabla\mathcal{L}(\theta^{(0)}) + \lambda^{\{K\}} \odot \xi\|_\infty + \|(\lambda^{\{K+1\}} - \lambda^{\{K\}}) \odot \xi\|_\infty$$

$$\overset{(ii)}{\leq} \omega_{\lambda^{\{K\}}}(\theta^{(0)}) + \|\lambda^{\{K+1\}} - \lambda^{\{K\}}\|_\infty \overset{(iii)}{\leq} \frac{\lambda_{\text{tgt}}}{8} + \frac{\lambda_{\text{tgt}}}{8} \leq \frac{\lambda_{\text{tgt}}}{4},$$

where $(i)$ is from triangle inequality, $(ii)$ is from the definition of the approximate KKT condition and $\xi$, and $(iii)$ is from $\omega_{\lambda^{\{K\}}}(\theta^{(0)}) = \omega_{\lambda^{\{K\}}}(\widehat{\theta}^{\{K\}}) \leq \frac{\lambda_{\text{tgt}}}{8}$ and $\|\lambda^{\{K+1\}} - \lambda^{\{K\}}\|_\infty \leq \frac{\lambda_{\text{tgt}}}{8}$.

For some $\xi^{(t)} = \mathrm{argmin}_{\xi \in \partial \|\theta^{(t)}\|_1} \|\nabla \mathcal{L}(\theta^{(t)}) + \lambda^{\{K+1\}} \odot \xi\|_\infty$, we have

$$\mathcal{F}_{\lambda^{\{K+1\}}}(\theta^*) \overset{(i)}{\geq} \mathcal{F}_{\lambda^{\{K+1\}}}(\theta^{(t)}) - (\nabla \mathcal{L}(\theta^{(t)}) + \lambda^{\{K+1\}} \odot \xi^{(t)})^\top (\theta^{(t)} - \theta^*)$$
$$\geq \mathcal{F}_{\lambda^{\{K+1\}}}(\theta^{(t)}) - \|\nabla \mathcal{L}(\theta^{(t)}) + \lambda^{\{K+1\}} \odot \xi^{(t)}\|_\infty \|\theta^{(t)} - \theta^*\|_1$$
$$\overset{(ii)}{\geq} \mathcal{F}_{\lambda^{\{K+1\}}}(\theta^{(t)}) - \frac{\lambda_{\mathrm{tgt}}}{4} \|\theta^{(t)} - \theta^*\|_1$$

where $(i)$ is from the convexity of $\mathcal{F}_{\lambda^{\{K+1\}}}$ and $(ii)$ is from the fact that for all $t \geq 0$, $\|\nabla \mathcal{L}(\theta^{(t)}) + \lambda^{\{K+1\}} \odot \xi^{(t)}\|_\infty \leq \frac{\lambda_{\mathrm{tgt}}}{4}$. This finishes the proof. □

**Lemma E.2** (Adapted from Fan et al. (2015)). *Suppose $\|\theta^{(t)}_{\mathcal{S}_\perp}\|_0 \leq \widetilde{s}$ and $\omega_{\lambda^{\{K\}}}(\theta^{(t)}) \leq \frac{\lambda_{\mathrm{tgt}}}{4}$. Then there exists a generic constant $c_1$ such that*

$$\|\theta^{(t)} - \theta^*\|_2 \leq \frac{c_1 \lambda_{\mathrm{tgt}} \sqrt{s^*}}{\rho^-_{s^*+2\widetilde{s}}}.$$

## F  Global Convergence Analysis

For notational convenience, we denote $\mathcal{F} = \mathcal{F}_\lambda$ and $\mathcal{R} = \mathcal{R}^{\ell_1}_\lambda$ in the sequel. We first provide an upper bound of the objective gap.

**Lemma F.1.** *Suppose the $\mathcal{F}(\theta) = \mathcal{R}(\theta) + \mathcal{L}(\theta)$ and $\mathcal{L}(\theta)$ satisfies the restricted Hessian smoothness property, namely, for any $\theta, h \in \mathbb{R}^d$*

$$\frac{d}{d\tau} \nabla^2 \mathcal{L}(\theta + \tau h)|_{\tau=0} \leq C \sqrt{h^\top \nabla^2 \mathcal{L}(\theta) h} \cdot \nabla^2 \mathcal{L}(\theta),$$

*for some constant $C$. Let $\Delta\theta$ be the search direction and let $\theta_+ = \theta + \tau \Delta\theta$ for some $\tau \in (0, 1]$. Then*

$$\mathcal{F}(\theta_+) \leq \mathcal{F}(\theta) + \left[-\tau + \mathcal{O}(\tau^2)\right] \|\Delta\theta\|_H^2.$$

*Proof.* From the convexity of $\mathcal{R}$, we have

$$\mathcal{F}(\theta_+) - \mathcal{F}(\theta) = \mathcal{L}(\theta_+) - \mathcal{L}(\theta) + \mathcal{R}(\theta_+) - \mathcal{R}(\theta)$$
$$\leq \mathcal{L}(\theta_+) - \mathcal{L}(\theta) + \tau \mathcal{R}(\theta + \Delta\theta) + (1-\tau)\mathcal{R}(\theta) - \mathcal{R}(\theta)$$
$$= \mathcal{L}(\theta_+) - \mathcal{L}(\theta) + \tau (\mathcal{R}(\theta + \Delta\theta) - \mathcal{R}(\theta))$$
$$= \nabla \mathcal{L}(\theta)^\top \cdot (\tau \Delta\theta) + \tau(\mathcal{R}(\theta + \Delta\theta) - \mathcal{R}(\theta)) + \tau \int_0^\tau (\Delta\theta)^\top \nabla^2 \mathcal{L}(\theta + \alpha \Delta\theta) \Delta\theta \, d\alpha.$$

By Lemma 6.7 and the restricted Hessian smoothness property, we obtain

$$\mathcal{F}(\theta_+) - \mathcal{F}(\theta) \leq -\tau \|\Delta\theta\|_{\nabla^2 \mathcal{L}(\theta)} + \tau \int_0^\tau (\Delta\theta)^\top \nabla^2 \mathcal{L}(\theta + \alpha \Delta\theta) \Delta\theta \, d\alpha$$
$$= -\tau \|\Delta\theta\|_{\nabla^2 \mathcal{L}(\theta)} + \tau \int_0^\tau d\alpha \int_0^\alpha dz \frac{d}{dz} (\Delta\theta)^\top \nabla^2 \mathcal{L}(\theta + z\Delta\theta) \Delta\theta + \tau \int_0^\tau d\alpha (\Delta\theta)^\top \nabla^2 \mathcal{L}(\theta) \Delta\theta$$
$$= \left(-\tau + \mathcal{O}(\tau^2)\right) \|\Delta\theta\|^2_{\nabla^2 \mathcal{L}(\theta)}.$$

□

Next, we show that $\Delta\theta \neq 0$ when $\theta$ have not attained the optimum.

**Lemma F.2.** *Suppose the $\mathcal{F}(\theta) = \mathcal{R}(\theta) + \mathcal{L}(\theta)$ has a unique minimizer, and $\mathcal{L}(\theta)$ satisfies the restricted Hessian smoothness property. Then $\Delta\theta^{(t)} = 0$ if and only if $\theta^{(t)} = \overline{\theta}$.*

*Proof.* Suppose $\Delta\theta$ is non-zero at $\overline{\theta}$. Lemma F.1 implies that for sufficiently small $0 < \tau \leq 1$,

$$\mathcal{F}(\overline{\theta} + \tau\Delta\theta^{(t)}) - \mathcal{F}(\overline{\theta}) \leq 0.$$

However $\mathcal{F}(\theta)$ is uniquely minimized at $\overline{\theta}$, which is a contradiction. Thus $\Delta\theta = 0$ at $\overline{\theta}$.

Now we consider the other direction. Suppose $\Delta\theta = 0$, then $\theta$ is a minimizer of $\mathcal{F}$. Thus for any direction $h$ and $\tau \in (0, 1]$, we obtain

$$\nabla\mathcal{L}(\theta)^\top (\tau h) + \frac{1}{2}\tau^2 h^\top H h + \mathcal{R}(\theta + \tau h) - \mathcal{R}(\theta) \geq 0.$$

Rearrange, we obtain

$$\mathcal{R}(\theta + \tau h) - \mathcal{R}(\theta) \geq -\tau\nabla\mathcal{L}(\theta)^\top h - \frac{1}{2}\tau^2 h^\top H h$$

Let $D\mathcal{F}(\theta, h)$ be the directional derivative of $\mathcal{F}$ at $\theta$ in the direction $h$, thus

$$\begin{aligned}
D\mathcal{F}(\theta, h) &= \lim_{\tau \to 0} \frac{\mathcal{F}(\theta + \tau h) - \mathcal{F}(\theta)}{\tau} \\
&= \lim_{\tau \to 0} \frac{\tau\nabla\mathcal{L}(\theta)^\top h + \mathcal{O}(\tau^2) + \mathcal{R}(\theta + \tau h) - \mathcal{R}(\theta)}{\tau} \\
&\geq \lim_{\tau \to 0} \frac{\tau\nabla\mathcal{L}(\theta)^\top h + \mathcal{O}(\tau^2) - \tau\nabla\mathcal{L}(\theta)^\top h - \frac{1}{2}\tau^2 h^\top H h}{\tau} = 0.
\end{aligned}$$

Since $\mathcal{F}$ is convex, then $\theta$ is the minimizer of $\mathcal{F}$. □

Then, we show the behavior of $\|\Delta\theta\|_H$ and $\mathcal{R}(\theta + \Delta\theta)$ when $\Delta\theta \neq 0$.

**Lemma F.3.** *Suppose at any point $\theta \in \mathbb{R}^d$, we have $\nabla\mathcal{L}(\theta) \in \mathrm{span}(\nabla^2 \mathcal{L}(\theta))$. If $\Delta\theta \neq 0$ then either*

$$\|\Delta\theta\|_H > 0 \quad or \quad \mathcal{R}(\theta + \Delta\theta) < \mathcal{R}(\theta).$$

*Proof.* Recall that $\Delta\theta$ is obtained by solving the following sub-problem,

$$\Delta\theta = \underset{\Delta\theta}{\mathrm{argmin}}\, \mathcal{R}(\theta + \Delta\theta) + \nabla\mathcal{L}(\theta)^\top \Delta\theta + \|\Delta\theta\|_H^2.$$

If $\|\Delta\theta\|_H = 0$ and $\Delta\theta \neq 0$, then

$$\Delta\theta \perp \mathrm{span}(H) \quad \text{and} \quad \nabla\mathcal{L}(\theta)^\top \Delta\theta = 0.$$

Thus

$$\mathcal{R}(\theta + \Delta\theta) < \mathcal{R}(\theta).$$

Notice that $\mathcal{R}(\theta + \Delta\theta) \neq \mathcal{R}(\theta)$, since otherwise $\Delta\theta = 0$ is a solution. □

Finally, we demonstrate the strict decrease of the objective in each proximal Newton step.

**Lemma F.4.** *Suppose at any point $\theta \in \mathbb{R}^d$, we have $\nabla \mathcal{L}(\theta) \in \text{span}\left(\nabla^2 \mathcal{L}(\theta)\right)$. If $\Delta \theta \neq 0$ then*

$$\mathcal{F}(\theta + \tau \Delta \theta) < \mathcal{F}(\theta),$$

*for small enough $\tau > 0$.*

*Proof.* By Lemma F.3, if $\Delta \theta \neq 0$, then either $\|\Delta \theta\|_H > 0$ or $\mathcal{R}(\theta + \Delta \theta) - \mathcal{R}(\theta) < 0$. If it is the first case, then by Lemma 6.7,

$$\gamma = \nabla \mathcal{L}(\theta)^\top \Delta \theta + \mathcal{R}(\theta + \Delta \theta) - \mathcal{R}(\theta) < -\|\Delta \theta\|_H < 0.$$

It is the second case, then $\nabla \mathcal{L}(\theta)^\top \Delta \theta = 0$ and

$$\gamma = \mathcal{R}(\theta + \Delta \theta) - \mathcal{R}(\theta) < 0.$$

Moreover, we have

$$\begin{aligned}
\mathcal{F}(\theta + \tau \Delta \theta) - \mathcal{F}(\theta) &= \mathcal{L}(\theta + \tau \Delta \theta) - \mathcal{L}(\theta) + \mathcal{R}(\theta + \tau \Delta \theta) - \mathcal{R}(\theta) \\
&\leq \tau \nabla \mathcal{L}(\theta)^\top \Delta \theta + \frac{\tau^2}{2} \Delta \theta^\top H \Delta \theta + \mathcal{O}(\tau^3) + \mathcal{R}(\theta + \tau \Delta \theta) - \mathcal{R}(\theta) \\
&\leq \tau \nabla \mathcal{L}(\theta)^\top \Delta \theta + \tau \mathcal{R}(\theta + \Delta \theta) + (1 - \tau) \mathcal{R}(\theta) - \mathcal{R}(\theta) + \frac{\tau^2}{2} \Delta \theta^\top H \Delta \theta + \mathcal{O}(\tau^3) \\
&= \tau(\gamma + \mathcal{O}(\tau)).
\end{aligned}$$

where the first inequality is from the restricted Hessian smoothness property. Thus $\mathcal{F}(\theta + \tau \Delta \theta) - \mathcal{F}(\theta) < 0$ for sufficiently small $\tau > 0$. $\square$

Since each step, the objective is strictly decreasing, thus the algorithm will eventually reach the minimum.